\pdfoutput=1
\documentclass[11pt]{article}

\usepackage[]{acl}

\usepackage{times}
\usepackage{latexsym}

\usepackage[T1]{fontenc}
\usepackage[utf8]{inputenc}

\usepackage{microtype}

\usepackage{amsfonts,amssymb}
\usepackage{mathrsfs}
\usepackage{multirow}
\usepackage{multicol}
\usepackage{float}
\usepackage{makecell}
 
\usepackage{arydshln}
\usepackage{ulem}
\usepackage{booktabs}
\usepackage{graphicx}
\usepackage{amsmath}
\usepackage{xspace}

\newcommand{\trp}[1]{\textit{\small[#1]}\xspace}
\newcommand{\trpsm}[1]{\textit{\small[#1]}\xspace}

\title{An Interpretable Neuro-Symbolic Reasoning Framework for\\ 
Task-Oriented Dialogue Generation}

\author{Shiquan Yang$^1$, Rui Zhang$^2$, Sarah Erfani$^1$, and Jey Han Lau$^1$\\
$^1$The University of Melbourne, $^2$www.ruizhang.info\\
$^1$\{shiquan@student., sarah.erfani@, laujh@\}unimelb.edu.au,
$^2$rayteam@yeah.net}

\begin{document}
\maketitle

\begin{abstract}
We study the interpretability issue of task-oriented dialogue systems in this paper. 
Previously, most neural-based task-oriented dialogue systems employ an
\textit{implicit} reasoning strategy that makes the model predictions uninterpretable to 
humans. To obtain a transparent reasoning process, we introduce neuro-symbolic to perform \textit{explicit} reasoning that justifies model decisions by reasoning chains. Since deriving reasoning chains requires multi-hop reasoning for task-oriented dialogues, existing neuro-symbolic approaches would induce error propagation due to the one-phase design. To overcome this, we propose a two-phase approach that consists of a hypothesis generator and a reasoner. We first obtain multiple hypotheses, i.e., potential operations to perform the desired task, through the hypothesis generator. Each hypothesis is then verified by the reasoner, and the valid one is selected to conduct the final prediction. The whole system is trained by exploiting raw textual dialogues without using any reasoning chain annotations. Experimental studies on two public benchmark datasets demonstrate that the proposed approach not only achieves better results, but also introduces an interpretable decision process. Code and data: \url{https://github.com/shiquanyang/NS-Dial}.
\end{abstract}

\section{Introduction}
\begin{figure}[!t]
\setlength{\belowcaptionskip}{-4mm}
    \centering
    \includegraphics[width=3.1in]{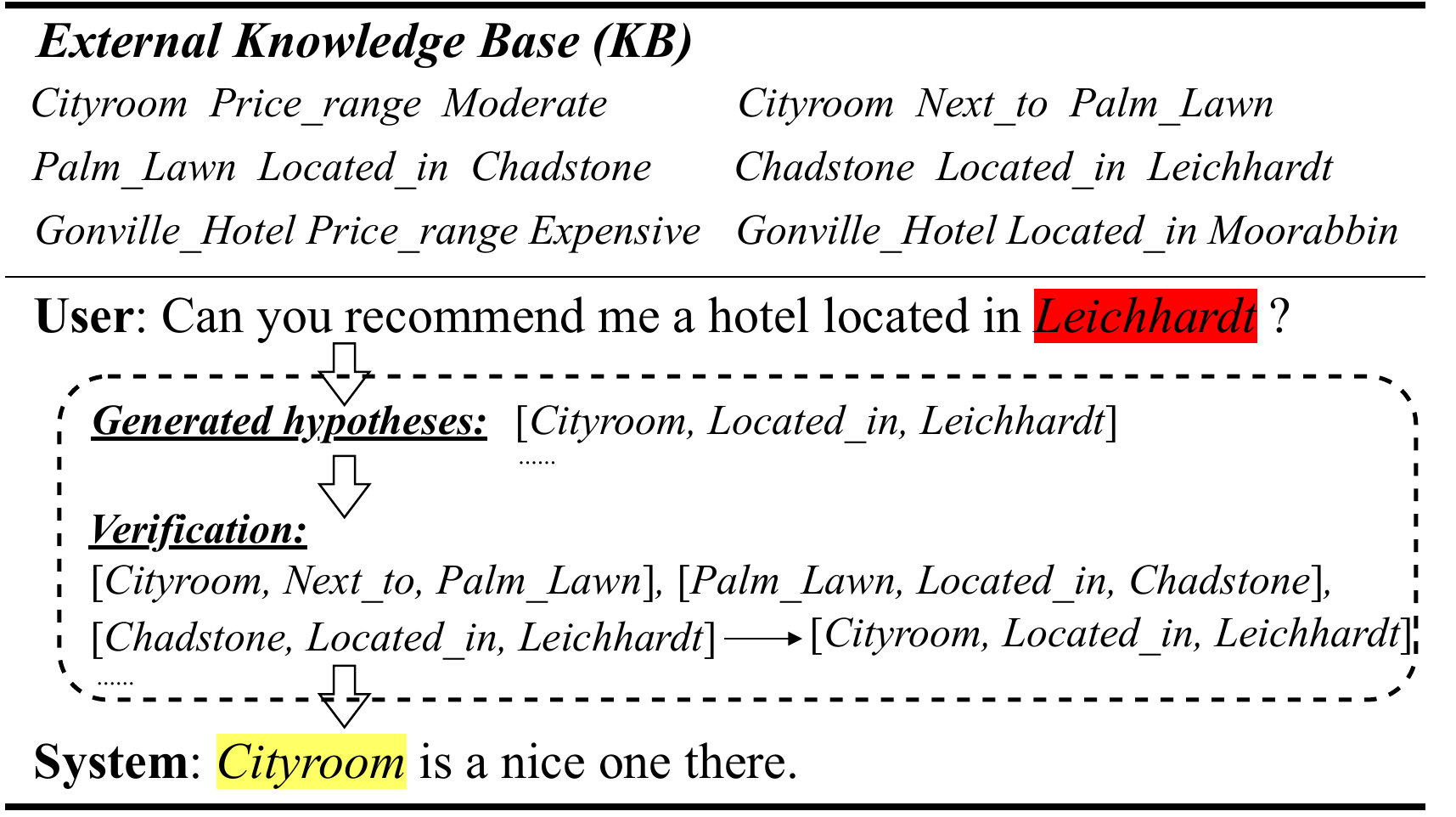}
    \caption{An example dialogue that incorporates external KB. The context entity (i.e., \textit{Leichhardt}) and answer entity (i.e., \textit{Cityroom}) are marked as {\textcolor{red}{Red}} and {\textcolor[RGB]{255,215,0}{Yellow}}, respectively. The triple containing the context entity and answer entity is not directly stored in KB and should be derived by a reasoning chain formed by multiple KB triplets.}
    \label{example_dialogue}
\end{figure}

Neural task-oriented dialogue systems have enjoyed a rapid progress recently \cite{peng2020soloist,hosseini2020simple,wu2020tod}, achieving strong empirical results on various benchmark datasets such as SMD \cite{eric2017key} and MultiWOZ \cite{budzianowski2018multiwoz}. However, most existing approaches suffer from the lack of explainability due to the black-box nature of neural networks \cite{doshi2017towards,lipton2018mythos, bommasani2021opportunities}, which may hurt the trustworthiness between the users and the system. For instance, in Figure \ref{example_dialogue}, a user is asking for a hotel recommendation at a given location. The system performs reasoning on a knowledge base (KB) and incorporates the correct entity in the response. However, when the system fails to provide the correct entities, it would be difficult for humans to trace back the issues and debug the errors due to its intrinsic \textit{implicit} reasoning nature. As a result, such system cannot be sufficiently trusted to be deployed in real-world products.

To achieve trustworthy dialogue reasoning, we aim to develop an interpretable KB reasoning as it's crucial for not only providing useful information (e.g., locations in Figure \ref{example_dialogue}) to users, but also essential for communicating options and selecting target entities. Without interpretability, it's difficult for users to readily trust the reasoning process and the returned entities.

% (e.g., rule-based expert systems)
To tackle this challenge, we present a novel \textbf{N}euro-\textbf{S}ymbolic 
\textbf{Dial}ogue framework (\textbf{NS-Dial}) which combines representation capacities of neural networks and explicit reasoning nature of symbolic approaches (e.g., rule-based expert systems). Existing neuro-symbolic approaches \cite{vedantam2019probabilistic,Chen2020Neural}
mostly employ a one-phase procedure where a tree-structured program composed of pre-defined human interpretable neural modules (e.g., attention and classification modules in Neural Module Networks \cite{andreas2016neural}) is generated to execute to obtain the final predictions. However, since the KB reasoning task involves a reasoning process spanning over multiple triplets
in a diverse and large-scale KB, only generating and following a single program (i.e., a reasoning chain formed by KB triplets) is prone to error propagation where a mistake in one step could lead to a failure of the subsequent reasoning process and may result in sub-optimal performances.

To address this, we propose a two-phase procedure to alleviate the effects of error propagation by first generating and then verifying multiple \emph{hypotheses}. Here, a hypothesis is in the form of a triplet containing an entity mentioned in dialogue context and an entity within KB, and their corresponding relation. The valid (i.e., correct) hypothesis is the one that contains the entity mentioned in the ground-truth response. Once we obtain multiple hypothesis candidates during the generation phase, we employ a reasoning engine for verifying those hypotheses. For instance in Figure \ref{example_dialogue}, given the user query ``\textit{Can you recommend 
me a hotel located in Leichhardt?}'', in order to find the valid hypothesis, the hypothesis generator obtains multiple candidates e.g.,\ 
\trp{Cityroom, Located\_in, Leichhardt} and \trp{Gonville\_Hotel, 
Located\_in, Leichhardt}. The reasoning engine will then construct
proof trees to verify them, e.g.,\ for the first hypothesis 
\trp{Cityroom, Located\_in, Leichhardt}, it can be verified with the 
following reasoning chain in the KB:
\trp{Cityroom, Next\_to, Palm\_Lawn} $\rightarrow$ \trp{Palm\_Lawn, 
Located\_in, Chadstone} $\rightarrow$ \trp{Chadstone, Located\_in, 
Leichhardt}. The whole framework is trained end-to-end using raw dialogues and thus does not require additional intermediate labels for either the hypothesis generation or verification modules.

To summarize, our contributions are as follows:
\begin{itemize}
    \item We introduce a novel neuro-symbolic framework for interpretable
    KB reasoning in task-oriented dialogue systems.
    \item We propose a two-phase ``generating-and-verifying'' approach 
    which generates multiple hypotheses and verifies them via reasoning chains
    to mitigate the error-propagation issue.
    \item We conduct extensive experimental studies on two benchmark 
    datasets to verify the effectiveness of our proposed model.
    By analyzing the generated hypotheses and the verifications, 
    we demonstrate our model's interpretability.

\end{itemize}

\section{Related Work}
\textbf{Task-Oriented Dialogue}\quad Traditionally, task-oriented dialogue systems are built via pipeline-based approaches where task-specific modules are designed separately and connected to generate system responses \cite{chen2016end,zhong2018global,wu2019transferable,chen2019semantically,huang-etal-2020-semi}. In another spectrum, many works have started to shift towards end-to-end approaches to reduce human efforts \cite{bordes2016endtoendtaskorientedlearning,lei-etal-2018-sequicity,mem2seq,moon-etal-2019-opendialkg,jung-etal-2020-attnio}. \citet{lei-etal-2018-sequicity} propose a two-stage sequence-to-sequence model to incorporate dialogue state tracking and response generation jointly in a single sequence-to-sequence architecture. \citet{zhang2020task} propose a domain-aware multi-decoder network (DAMD) to combine belief state tracking, action prediction and response generation in a single neural architecture. 
% \citet{lei-etal-2018-sequicity} propose a two-stage sequence-to-sequence model to incorporate dialogue state tracking and response generation jointly in a single sequence-to-sequence architecture. \citet{zhang2020task} propose a domain-aware multi-decoder network (DAMD) to combine belief state tracking, action prediction and response generation in a single neural architecture. 
Most recently, the success of large-scale pre-trained language models (e.g., BERT, GPT-2) \cite{devlin2018bert,radford2019language} has spurred a lot of recent dialogue studies starting to explore large-scale pre-trained language model for dialogues \cite{wolf2019transfertransfo,zhang2019dialogpt}. In task-oriented dialogue, \citet{budzianowski2019hello} use GPT-2 to fine-tune on MultiWOZ dataset for dialogue response generation. \citet{peng2020soloist} and \citet{hosseini2020simple} employed a single unified GPT-2 model jointly trained for belief state prediction, system action and response generation in a multi-task fashion. However, most existing approaches cannot explain why the model makes a specific decision in a human understandable way. We aim to address this limitation and introduce interpretability for dialogue reasoning in this study.

\noindent\textbf{Neuro-Symbolic Reasoning}\quad Neuro-Symbolic reasoning has attracted a lot of research attentions recently due to its advantage of exploiting the representational power of neural networks and the compositionality of symbolic reasoning for more robust and interpretable models \cite{andreas2016neural,hu2017learning,hudson2018compositional,vedantam2019probabilistic,chen2019neural,vedantam2019probabilistic,VANKRIEKEN2022103602}. The main difference between neuro-symbolic vs. pure neural networks lies in how the former combines basic rules or modules to model complex functions. \citet{rocktaschel2017end} propose a neuro-symbolic model that can jointly learn sub-symbolic representations and interpretable rules from data via standard back-propagation. In visual QA, \citet{andreas2016neural} propose neural module networks to compose a chain of differentiable modules wherein each module implements an operator from a latent program. \citet{yi2018neural} propose to discover symbolic program trace from the input question and then execute the program on the structured representation of the image for visual question answering. However, these approaches cannot be easily adapted to task-oriented dialogues due to the error propagation issue caused by multi-hop reasoning on large-scale KBs. Thus, we aim to bridge this gap by developing a neuro-symbolic approach for improving task-oriented dialogues.

\begin{figure*}[!ht]
    \centering
    \includegraphics[width=6.4in]{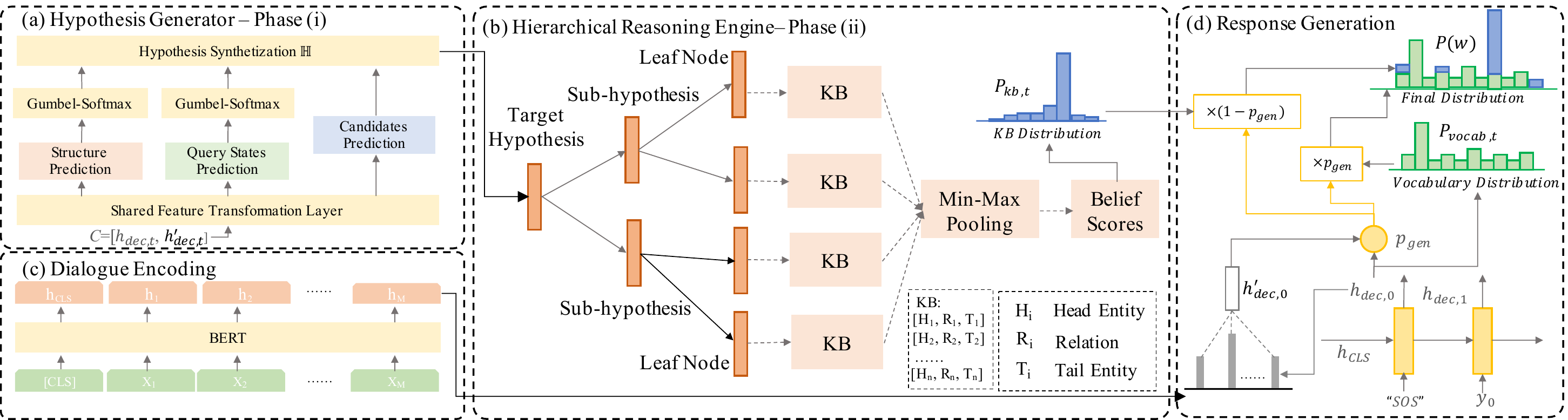}
    \caption{Illustration of the overall architecture: (a) hypothesis generator generating a set of synthesized hypotheses; (b) reasoning engine used to verify the generated hypotheses; (c) dialogue encoding; (d) response generation.}
    \label{framework}
\end{figure*}

\section{Preliminary}
In this work, we focus on the problem of task-oriented dialogue response generation with KBs. Formally, given the dialogue history $X$ and knowledge base $B$, our goal is to generate the system responses $Y$ word-by-word. The probability of the generated responses can be written as:

\begin{equation}
    \small
    p(Y|X,B) = \prod_{t=1}^{n} p(y_{t}|X,B,y_{1},y_{2},...,y_{t-1})
\end{equation}

% \vspace{-1mm}
\noindent where $y_{t}$ is the t-th token in the response $Y$. The overall architecture is
shown in Figure \ref{framework}. We start by introducing the standard modules in our system and then explain the two novel modules afterward.

\subsection{Dialogue Encoding}
We employ pre-trained language model BERT 
\cite{devlin2019bert} as the backbone to obtain the distributed 
representations for each token in the dialogue history. 
Specifically, we add a $[CLS]$ token at the start of the dialogue 
history to represent the overall semantics of the dialogue. The hidden states $H_{enc}$ = ($h_{CLS},h_{1},...,h_{M}$) for all the input tokens $X$ = ($[CLS],x_{1},...,x_{M}$) are computed using:

% \vspace{-2mm}
\begin{equation}
    \small
    {H_{enc}} = \textup{BERT}_{\textup{\scriptsize enc}}(\phi^{\scriptsize emb}({X}))
\end{equation}

% \vspace{-1mm}
\noindent where $M$ is the number of tokens in the dialogue history, $\phi^{\scriptsize emb}$ is the embedding layer of BERT.

\subsection{Response Generation}
To generate the system response, we first utilize a linear layer to project $H_{enc}$ 
to $H_{enc}^{'}$ = ($h_{CLS}^{'},h_{1}^{'},...,h_{M}^{'}$) that are in 
the same space of the decoder. We initialize the decoder with 
$h_{CLS}^{'}$. During decoding timestep \textit{t}, the model utilizes 
the hidden state $h_{dec,t}$ to attend $H_{enc}^{'}$ to 
obtain an attentive representation $h_{dec,t}^{'}$ via standard attention mechanism. 
We then concatenate $h_{dec,t}$ and $h_{dec,t}^{'}$ to form a context vector $C$ and project 
it into the vocabulary space $\mathcal{V}$:

\vspace{-4mm}
\begin{small}
\begin{align}
    C &= [h_{dec,t},h_{dec,t}^{'}]  \label{eqn:c}\\
    P_{vocab,t} &= \textup{Softmax}(U_1 C)
\end{align}
\end{small}

\vspace{-5mm}
\noindent where $U_1$ is a learnable linear layer, $P_{vocab,t}$ is the vocabulary distribution for generating the token $y_t$.

Next, we aim to estimate the KB distribution $P_{kb,t}$, i.e., the probability distribution of entities in the KB, in an interpretable way and fuse $P_{vocab,t}$ and $P_{kb,t}$ for generating the final output tokens.
We follow \citet{see-etal-2017-get} and employ a soft-switch mechanism 
to fuse $P_{vocab,t}$ and $P_{kb,t}$ to generate output token $y_t$.  
Specifically, the generation probability $p_{gen}$ $\in$ [0,1] is computed from the attentive representation 
$h_{dec,t}^{'}$ and the hidden state $h_{dec,t}$:

% \vspace{-2mm}
\begin{equation}
    \small
    p_{gen} = \sigma(U_2([h_{dec,t}^{'},h_{dec,t}]))
\end{equation}

% \vspace{-2mm}
\noindent where $\sigma$ is sigmoid function, $U_2$ is a linear layer. The output token $y_t$ is generated by greedy sampling from the probability distribution $P(w)$:

% \vspace{-2mm}
\begin{equation}
    \small
    P(w) = p_{gen}P_{vocab,t}+(1-p_{gen})P_{kb,t}
    \label{kb_dis}
\end{equation}

We next describe how to obtain the KB distribution $P_{kb,t}$ in details using the two novel modules we proposed, i.e., hypothesis generator and hierarchical reasoning engine.

\section{Neuro-Symbolic Reasoning For Task-Oriented Dialogue}\label{approach}
To compute the KB distribution $P_{kb,t}$, we present two novel modules: 
hypothesis generator (HG) and hierarchical reasoning engine (HRE). 
We take the context vector $C$ (Equation \ref{eqn:c}) as the input of HG 
module and generate K hypotheses $\mathbb{H}$, each of which are then 
fed into the HRE module to generate the logical reasoning chains and 
their belief scores.  The estimated belief scores are then served as 
$P_{kb,t}$, giving us a distribution over the entities in the KB. Next, 
we describe how each component works in detail and explain how they
interact with each other for generating $P_{kb,t}$.

\subsection{Hypothesis Generator}\label{hypothesis_generator}
Let a hypothesis be a 3-tuple of the form 
``$[H, R, T]$'', where $H$ and 
$T$ are the head and tail entities, and $R$ is the 
relation between entities. In this paper, we are interested in three 
types of hypotheses including the H-Hypothesis, T-Hypothesis, and 
R-Hypothesis. The H-Hypothesis is the structure where the tail entity 
$T$ and relation $R$ are inferred from the context and 
the head entity $H$ is unknown (which needs to be answered 
using the KB), and it takes the form 
``$[\triangleright, R, T]$''. In a similar vein, the  
T-Hypothesis and R-Hypothesis have unknown tail entity $T$ and 
relation $R$, respectively. The goal of 
the Hypothesis Generator module is to generate hypotheses in this 
triple format which will later be verified by the Hierarchical Reasoning 
Engine.

Intuitively, a hypothesis can be determined by its content and structure. The 
structure indicates the template form of the hypothesis while the 
content fills up the template. For instance, the H-Hypothesis has its 
template form of ``$[\triangleright, R, T]$'' and the 
content that needs to be realised includes candidate entities (i.e., 
``$\triangleright$''), and query states (i.e., the tail ``$T$'' and relation 
entities ``$R$''). To this end, we employ a divide-and-conquer 
strategy to jointly learn three sub-components: structure prediction, query states prediction, 
and candidates prediction. Next, we describe each sub-component in details.

\noindent\textbf{Structure Prediction (SP)}\quad The goal of the 
structure prediction module is to determine the structure of the 
hypothesis (i.e.,\ H/T/R-Hypothesis) based on the context. For example in 
Figure \ref{example_dialogue}, one might expect an H-Hypothesis at 
timestep $0$. Specifically, SP uses a shared-private architecture to 
predict the hypothesis type. It first takes the context vector $C$ (Equation \ref{eqn:c}) as input and 
utilizes a shared transformation layer between all the three sub-components to 
learn task-agnostic feature $h_{\scriptsize share}$:

% \vspace{-2mm}
\begin{equation}
    \small
    h_{share} = W_2(\textup{LeakyReLU}(W_1C))
\end{equation}

% \vspace{-1mm}
\noindent where $W_1$ and $W_2$ are learnable parameters (shared by the structure 
prediction, query states prediction and candidate prediction components) 
and \textup{LeakyReLU} is the activation function.

The shared layer can be parameterised with complicated neural 
architectures.  However, to keep our model simple, we use linear 
layers which we found to perform well in our experiments.  SP next uses a private layer on top of the shared layer to 
learn task-specific features for structure prediction:

% \vspace{-2mm}
\begin{equation}
    \small
   h_{\scriptsize private}^{sp} = W_4(\textup{LeakyReLU}(W_3h_{\scriptsize share}))
\end{equation}

% \vspace{-1mm}
\noindent where $W_3$ and $W_4$ are learnable parameters. For ease of presentation, we define the private feature transformation function as:

% \vspace{-2mm}
\begin{equation}
    \small
    \mathcal{F}^{\star}: h_{\scriptsize share} \rightarrow h_{\scriptsize private}^{\star}
\end{equation}

% \vspace{-1mm}
\noindent where $\star$ denotes any of the three sub-components. To 
obtain the predicted hypothesis structure, a straightforward approach is to 
apply softmax on $h_{private}^{sp}$. However, this will break the 
differentiability of the overall architecture since we perform sampling 
on the outcome and pass it to the neural networks. To avoid this, we 
utilize the Gumbel-Softmax trick \cite{jang2016categorical} over 
$h_{private}^{sp}$ to get the sampled structure type:

% \vspace{-2mm}
\begin{equation}
    \small
    \textup{I}_{sp} = \textup{Gumbel-Softmax}(h_{\scriptsize private}^{sp}) \in \mathbb{R}^{3}
\end{equation}

% \vspace{-1mm}
\noindent where $\textup{I}_{sp}$ is a one-hot vector and the index of one element 
can be viewed as the predicted structure. In this paper, we define 0 as H-Hypothesis, 1 as T-Hypothesis and 
2 as R-Hypothesis.

% The query states can be abstracted as two tokens for each type of the hypothesis defined above.
\noindent\textbf{Query States Prediction (QSP)}\quad
Query states are the tokens in hypothesis that need to be 
inferred from the dialogue history. For example, one might want to infer 
relation $R$=\textit{Located\_in} and tail 
$T$=\textit{Leichhardt} based on the history in Figure 
\ref{example_dialogue}.  Therefore, the goal of the query states 
prediction is to estimate the state information (e.g., $T$ 
and $R$ in H-Hypothesis) of hypothesis.  
Specifically, QSP takes the shared feature $h_{share}$ as the input and 
next applies the private feature transformation function followed by 
Gumbel-Softmax to obtain the state tokens of hypothesis using:

% \vspace{-4mm}
\begin{small}
\begin{align}
    h_{private}^{qsp,k} &= \mathcal{F}^{qsp,k}(h_{\scriptsize share}) \\
    \textup{I}_{qsp}^{k} &= \textup{Gumbel-Softmax}(h_{private}^{qsp,k}) \in \mathbb{R}^{n}
\end{align}
\end{small}

% \vspace{-4mm}
\noindent where $n$ is the number of tokens (entities and relations) in the KB, $k$ $\in$ \{0,1\},\ $\textup{I}_{qsp}^{0}$ and $\textup{I}_{qsp}^{1}$ are two one-hot vectors where their corresponding tokens in KB serve as the state tokens of the hypothesis.

\noindent\textbf{Candidates Prediction (CP)}\quad
To generate the final hypotheses, we need multiple candidates to instantiate 
the structure of the hypothesis except the state tokens, e.g.,\   
\textit{Cityroom} or \textit{Gonville\_Hotel} as candidate head entities 
$H$ in Figure \ref{example_dialogue}. To this end, we utilize 
an embedding layer $\phi^{emb}_{cp}$ to convert all the tokens in the KB 
to vector representations. We then compute a probability distribution 
over all the KB tokens using:

% \vspace{-2mm}
\begin{equation}
    \small
    \textup{P}_{i} = \textup{Sigmoid}(\phi^{emb}_{cp}(K_i)\odot h_{share})
    \label{kb_distribution}
\end{equation}

% \vspace{-1mm}
\noindent where $K_i$ is the $i$-th token in KB, $\phi^{emb}_{cp}$ is the embedding layer of CP, $\textup{P}_{i}$ is the probability of the $i$-th token to be candidate, $\odot$ denotes inner-product. We use sigmoid instead of softmax as we find that softmax distribution is too ``sharp'' making the probability between different tokens are hard to differentiate for sampling multiple reasonable candidates.

\noindent\textbf{Hypothesis Synthesizing}\quad
The final hypotheses $\mathbb{H}$ are composed by combining the outputs of the three sub-components as follows: (i) We generate the hypothesis template according to 
the predicted structure type. For example, if SP predicts a 
structure type 0 which denotes H-Hypothesis, the model will form a 
template of ``$[\triangleright, R, T]$''; (ii) We next instantiate the state tokens in the hypothesis sequentially by using the outputs of QSP module. For example, if 
the output tokens of QSP are ``$\textit{Located\_in}$'' ($k$=0) and 
``$\textit{Leichhardt}$'' ($k$=1), the hypothesis will become 
$[\triangleright,\textit{Located\_in},\textit{Leichhardt}]$; (iii) Finally, we instantiate the candidate (i.e., $\triangleright$) 
with the top-$K$ ($K=$5 in our best-performing version) entities 
selected from $\textup{P}$. If the top-2 highest probability tokens are 
\textit{Cityroom} and \textit{Gonville\_Hotel}, the model will 
instantiate two hypotheses \trpsm{Cityroom, Located\_in, Leichhardt}, 
\trpsm{Gonville\_Hotel, Located\_in, Leichhardt}.

\subsection{Hierarchical Reasoning Engine}\label{hre}
With the hypotheses generated by HG module, we next aim to verify them via 
logical reasoning chains. Inspired by Neural Theorem Provers \cite{ntps},  we develop 
chain-like logical reasoning with following format:

% \vspace{-6mm}
\begin{equation}
    \small
    \alpha,\ (H, R, T) \leftarrow (H, R_n, Z_n) \wedge \cdots \wedge (Z_1, R_1, T)
    \label{logical_reasoning}
\end{equation}

% \vspace{-1mm}
\noindent where $\alpha$ is a weight indicating the 
\textit{belief} of the model on the target hypothesis 
$[H, R, T]$, and the right part of the arrow 
is the reasoning chain used to prove that hypothesis, and 
$R_i$ and $Z_i$ are relations and entities from the 
KB. The goal is to find the proof chain and the {confidence} 
$\alpha$ for a given hypothesis. To this end, we introduce a 
neural-network based hierarchical reasoning engine (HRE) that learns to 
conduct chain-like logical reasoning. At a high level, HRE recursively 
generates multiple levels of sub-hypotheses using neural networks that 
form a tree structure as shown in Figure \ref{framework}. Next, we 
describe how this module works in details.

The module takes the output hypotheses from the HG module as input. Each hypothesis serves as
one target hypothesis. To generate the reasoning chain in Equation \ref{logical_reasoning}, 
the module first finds sub-hypotheses of the same format as the target in the hypothesis space. The sub-hypotheses can be viewed as the intermediate reasoning results 
to prove the target. One straightforward approach is to use neural networks 
to predict all the tokens in the sub-hypotheses (2 heads, 2 tails and 2 
relations). However, this can lead to extremely large search space of 
triples and is inefficient. Intuitively,
sub-hypotheses inherit from the target hypothesis and sub-hypotheses 
themselves are connected by bridge entities. For example, 
\trp{Uber,office\_in,USA} can be verified by two sub-hypotheses 
\trp{Uber,office\_in,Seattle} and \trp{Seattle,a\_city\_of,USA}, 
\textit{Uber} and \textit{USA} are inherited from the 
target and \textit{Seattle} is the bridge entity between sub-hypotheses.  
Motivated by this, we propose to reduce the triple search complexity by 
constraining the sub-hypotheses. Specifically, 
given target $[H, R, T]$, we generate 
sub-hypotheses of the format 
$[H, R_1, Z]$,$[Z, R_2, T]$, 
where $Z$ is the bridge entity, $R_1$ and 
$R_2$ are relations to be predicted. 
Therefore, the goal of the neural networks has been reduced to predict 
three tokens (2 relations and 1 bridge entity). Formally, HRE predicts 
the vector representation of bridge entity as follows:

% \vspace{-4mm}
\begin{small}
\begin{align}
    h_{H},h_{R},h_{T} &= \phi^{emb}_{cp}(H),\phi^{emb}_{cp}(R),\phi^{emb}_{cp}(T)
    \label{embed_layer} \\
    h_{Z} &= W_{6}(\textup{LeakyReLU}(W_5[h_{H},h_{R},h_{T}]))
    \label{linear_layer}
\end{align}
\end{small}

% \vspace{-5mm}
\noindent where $[h_{H},h_{R},h_{T}]$ are the concatenation of the representations of tokens in target hypothesis, $h_{Z}$ is the vector representation of bridge entity $Z$. The prediction of $h_{R_1}$ and $h_{R_2}$ uses the same architecture in Equation \ref{linear_layer} and the difference is that they use different linear layers for the feature transformation. Note that $h_{Z}$ denotes a KB token in the embedding space. We can decode the token by finding the nearest KB token to $h_{Z}$ in vector space. More details on the token decoding can be found in Appendix A. Upon obtaining $h_{Z},h_{R_1},h_{R_2}$, the module generates the two sub-hypotheses in vector representations. Next, the module iteratively takes each of the generated sub-hypothesis as input and extend the proof process by generating next-level 
sub-hypotheses in a depth-first manner until the maximum depth $D$ has been reached.

\noindent\textbf{Belief Score}\quad
To model confidence in different reasoning chains, we 
further measure the semantic similarities between each triple of the 
leaf node and triples in the KB, and compute the belief score $\alpha_m$ of the m-th hypothesis $\mathbb{H}_m$:

% \vspace{-2mm}
\begin{equation}
    \small
    \alpha_{m} = \min_{\forall i \in U}\ \max_{\forall j \in V}\ e^{-d_j(Leaf_i, KB_j)}
\end{equation}

% \vspace{-1mm}
\noindent where $Leaf_i$ is the representation (concatenation of $H,R,T$) of the 
i-th leaf node in the proof tree (DFS manner), $KB_j$ is the 
representation of the j-th triple in KB, $U$=[0,...,$u$-1], 
$V$=[0,...,$v$-1] where $u$ and $v$ are the number of leaf nodes and KB 
triples correspondingly, $d$ is the distance metric. In general, any 
distance function can be applied and we adopt Euclidean distance in our 
implementation since we found that it worked well in our 
experiments. All the triples in the 
leaf nodes form the reasoning chain for the input hypothesis as in 
Equation \ref{logical_reasoning}. The hypotheses $\mathbb{H}$ coupled 
with the belief $\alpha$ form our KB distribution $P_{kb,t}$. More details can be found
in Appendix B. Intuitively, the belief score can be viewed as the likelihood of the hypothesis contains the correct entity. If the hypothesis is valid (i.e., contains the correct answer
entity), it should have a high likelihood and thus encourage to generate more proper reasoning chains based on the triples stored in the KB.

\noindent \textbf{Training}\quad
We apply two loss functions to train the whole architecture end-to-end. The first loss function $\mathcal{L}_{gen}$ is for the final output. We use a cross-entropy loss over the ground-truth token and the generated token from the final distribution $P(w)$. The second loss $\mathcal{L}_{cp}$ is for the candidates prediction (CP) module in the hypotheses generator. We apply binary cross-entropy loss over the output distribution for each KB token (Equation \ref{kb_distribution}) and their corresponding labels. The labels for each KB token are computed as follows:

% \vspace{-2mm}
\begin{equation}
\small
Label_i = \left\{
             \begin{array}{ll}
             1, & K_i = y_t  \\
             0, & K_i\neq y_t \\
             \end{array}
\right.
\end{equation}

% \vspace{-1mm}
\noindent where $K_i$ is the $i$-th token in the KB and $y_t$ is the ground-truth output at timestep t. The final loss $\mathcal{L}$ is calculated by:

% \vspace{-2mm}
\begin{equation}
    \mathcal{L} = \gamma_g*\mathcal{L}_{gen}+\gamma_c*\mathcal{L}_{cp}
\end{equation}

% \vspace{-1mm}
\noindent where $\gamma_g$ and $\gamma_c$ are hyper-parameters and we set them to 1 in our experiments.

\begin{table}
    \centering
    \scriptsize
    \scalebox{1.00}{
    \begin{tabular}{ccccc}
         \toprule[1pt]
         \textbf{Dataset} & \textbf{Domains} & \textbf{Train} & \textbf{Dev} & \textbf{Test}  \\
         \midrule
         {SMD} & Navigate,Weather,Schedule & 2425 & 302 & 304 \\
         \midrule
         {MultiWOZ 2.1} & Restaurant,Attraction,Hotel & 1839 & 117 & 141 \\
         \bottomrule[1pt]
    \end{tabular}}
    \caption{Statistics of {SMD} and {MultiWOZ 2.1}.}
    \label{dataset_statistics}
\end{table}

\begin{table*}[!t]
% \setlength{\belowcaptionskip}{-3mm}
    % \small
    \centering
    \scalebox{0.88}{
    \begin{tabular}{rcccccccccc}
        \toprule
         & \multicolumn{5}{c}{{SMD}} & \multicolumn{5}{c}{{MultiWOZ 2.1}} \\
         \midrule
         \multirow{2}{*}{{{Model}}} & \multirow{2}{*}{{{BLEU}}} & \multirow{2}{*}{{{F1}}} & {{Navigate}} & {{Weather}} & {{Calendar}} & \multirow{2}{*}{{{BLEU}}} & \multirow{2}{*}{{{F1}}} & {{Restaurant}} & {{Attraction}} & {{Hotel}} \\
         & & & {{F1}} & {{F1}}  & {{F1}}  & & & {{F1}}  & {{F1}}  & {{F1}}  \\
         \midrule
         {Mem2Seq} & 12.6 & {33.4} & {20.0} & {32.8} & {49.3} & {6.6} & {21.6} & {22.4} & {22.0} & {21.0} \\
         {GLMP} & {13.9} & {60.7} & {54.6} & {56.5} & {72.5} & {6.9} & {32.4} & {38.4} & {24.4} & {28.1} \\
         {GraphDialog} & {14.2} & {61.1} & {56.4} & {56.9} & {72.1} & {6.7} & {34.1} & {39.2} & {27.8} & {29.6} \\
         {DF-Net} & {14.4} & {62.7} & {57.9} & {57.6} & {73.1} & {9.4} & {35.1} & {40.9} & {28.1} & {30.6} \\
         \midrule
         {Ours (D=1)} & {14.9}  & {63.8} & {60.1} & {58.7} & {75.0} & {9.7} & {36.5} & {42.0} & {29.7} & {32.8} \\
         %\cdashline{1-11}[3pt/3pt]
         {Ours (D=3)} & ${\textbf{15.6}}^{*}$  & ${\textbf{64.5}}^{*}$ & ${\textbf{60.3}}^{*}$ & ${\textbf{59.2}}^{*}$ & ${\textbf{75.6}}^{*}$ & ${\textbf{10.6}}^{*}$ & ${\textbf{37.2}}^{*}$ & ${\textbf{42.6}}^{*}$ & ${\textbf{30.6}}^{*}$ & ${\textbf{33.7}}^{*}$ \\
         %\cdashline{1-11}[3pt/3pt]
         {Ours (D=5)} & {14.5}  & {63.5} & {59.4} & {57.9} & {74.8} & {9.3} & {36.2} & {41.7} & {28.8} & {31.5} \\
        %  \midrule
        %  {Ours (w/o MT)} & {14.8} & {63.7} & {58.7} & {58.6} & {74.1} & {9.5} & {36.4} & {41.5} & {29.1} & {32.4} \\
         \bottomrule
    \end{tabular}}
    \caption{Main results. D denotes the maximum depth of HRE module. We run each experiment 5 times with different random seeds and report the average results. * denotes that the improvement of our framework over all baselines are statistically significant with $p$\ \textless\ 0.05 under t-test. Following \citet{qin-etal-2020-dynamic}, we report \textit{Navigate, Weather, Calendar} on SMD and \textit{Restaurant, Attraction, Hotel} on MultiWOZ for per-domain results.}
    \label{main_results}
\end{table*}
%  Ours (w/o MT) denotes our model variant that use private feature transformation layers for sub-components in HG module without shared layers. 

\section{Experiments}
\subsection{Datasets}
To evaluate the effectiveness and demonstrate the interpretability of our proposed approach,
we conduct experiments on two public benchmark datasets for task-oriented dialogue in this paper, {SMD}
\cite{eric2017key} and {MultiWOZ 2.1} \cite{budzianowski2018multiwoz}. We 
use the partitions created by \citet{eric2017key,mem2seq} and 
\citet{qin-etal-2020-dynamic} for {{SMD}} and {MultiWOZ}, respectively.
Statistics of the datasets are presented in Table \ref{dataset_statistics}.
In the Appendix E, we present several additional results on a large-scale synthetic
dataset to demonstrate our model's multi-hop reasoning capability under complex
KB reasoning scenarios.

\subsection{Baselines}
We compare our model with the following state-of-the-art baselines on KB reasoning in task-oriented dialogues: (1)\ 
{{Mem2Seq}} \cite{mem2seq}:  employs memory networks to store the 
KB and combine pointer mechanism to either generate tokens from 
vocabulary or copy from memory; (2)\ {{GLMP}} \cite{wu2019global}: 
uses a global-to-local pointer mechanism to query the KB during 
decoding; (3)\ {{DF-Net}} \cite{qin-etal-2020-dynamic}: employs 
shared-private architecture to capture both domain-specific and 
domain-general knowledge to improve the model transferability; (4)\ 
{{GraphDialog}} \cite{yang-etal-2020-graphdialog}: incorporates graph structural information obtained from
sentence dependency parsing results for improving KB reasoning accuracy and response generation quality. Detailed experimental settings are included in Appendix C.

\subsection{Main Results}
Following prior work \cite{eric2017key,mem2seq,wu2019global}, we adopt 
the \textit{BLEU} and \textit{Entity F1} metrics to evaluate the 
performance of our framework. The results on the two datasets are shown 
in Table \ref{main_results}. As we can see, our framework
consistently outperforms all the previous state-of-the-art baselines on 
all datasets across both metrics. Specifically, on {MultiWOZ} dataset, 
our model achieves more than 2\% absolute improvement in Entity F1 and 
1.2\% improvement in BLEU over baselines. The improvement in Entity 
F1 indicates that our model enhances KB reasoning, while the increase 
in BLEU suggests that the quality of the generated responses has been
improved. The same trend has also been observed on {SMD} dataset. This 
indicates the effectiveness of our proposed framework
for task-oriented dialogue generation.

\subsection{Model Interpretability}
\begin{figure*}[!t]
    \centering
    \includegraphics[width=6.5in]{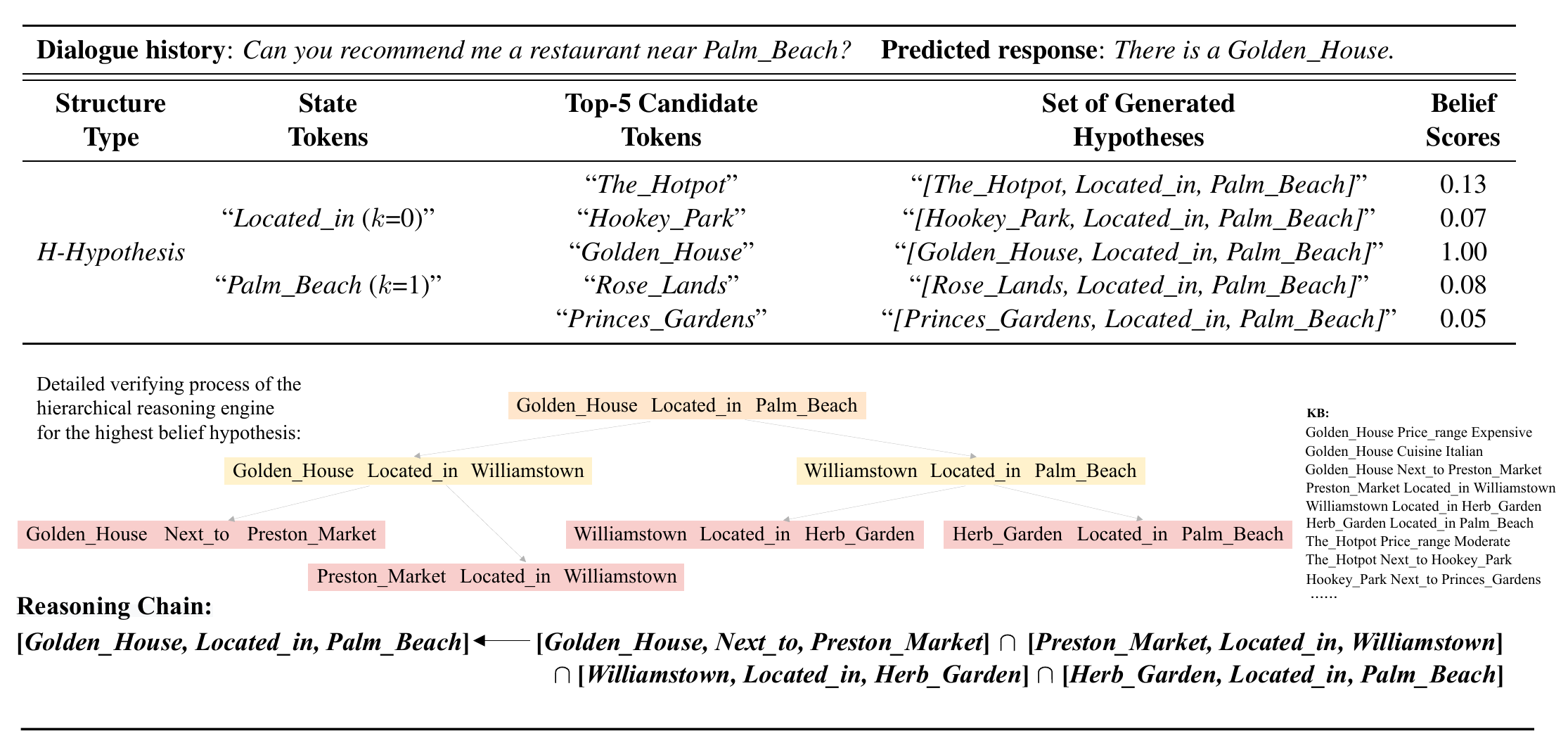}
    \caption{Example of inner workings of the hypothesis generator and hierarchical reasoning engine for generating \textit{Golden\_House} in the response given dialogue history \textit{Can you recommend me a restaurant near Palm\_Beach?}. Our model has performed a 4-hop reasoning to verify the target hypothesis \trp{Golden\_House, Located\_in, Palm\_Beach}.}
    \label{example_dialogue_2}
\end{figure*}

To demonstrate our framworks's interpretability, we investigate the inner workings of our 
framework.  As shown in Figure \ref{example_dialogue_2}, given the dialogue 
history ``\textit{Can you recommend me a restaurant near Palm\_Beach?}'', the generated 
response is ``\textit{There is a Golden\_House.}''. During the {3rd} timestep, 
our model has successfully predicted an appropriate
\textit{H-Hypothesis} with \textit{Located\_in} and \textit{Palm\_Beach} 
as its state tokens. Our model further instantiates five concrete 
hypotheses and computes their belief scores leveraging the reasoning 
engine, respectively. As we can see from the table, our model successfully generates five reasonable 
hypotheses and scores them correctly (with highest score for the 
oracle KB entity \textit{Golden\_House}). The proof process for the highest
score hypothesis is shown in Figure \ref{example_dialogue_2}. 
The verification procedure generated by the HRE module has a 
depth of 3 and the reasoning chaining used to verify the target 
hypothesis is: \trp{Golden\_House, Next\_to, Preston\_Market} $\rightarrow$ 
\trp{Preston\_Market, Located\_in, Williamstown} $\rightarrow$ \trp{Williamstown, 
Located\_in, Herb\_Garden} $\rightarrow$ \trp{Herb\_Garden, 
Located\_in, Palm\_Beach}. This indicates that our framework has successfully 
utilized the KB information to support the reasoning process \textit{explicitly} to reach a correct 
conclusion. More examples and error analyses can be found in the Appendix (Appendix E.4 and F).

\subsection{Ablation Study}
We ablate each component in our framework to study their effectiveness
on both datasets. The results are shown in Table \ref{ablation}.  
Specifically, 1) w/o HRE denotes that we simply use the 
probability in candidates prediction (CP) module (Equation \ref{kb_distribution}) as the KB distribution without using the scores from the reasoning engine. 2) w/o BERT denotes 
that we use standard GRU as encoder instead of BERT. 3) w/o Soft-switch 
denotes that we simply sum the KB distribution and vocabulary distribution without using a soft gate. As we can see from the table, all the individual components have notably 
contributed to the overall performance of our framework. Specifically, 
when removing HRE module, the performance has decreased substantially 
(more than 5\% absolute drop), which confirms that the effectiveness of the proposed 
hierarchical reasoner module.

\begin{table}
    \centering
    % \small
    \scalebox{0.90}{
    \begin{tabular}{lcccc}
         \toprule
         & \multicolumn{2}{c}{{SMD}} & \multicolumn{2}{c}{{MultiWOZ 2.1}} \\
         \midrule
         {{Model}} & {{F1}} {(\%)} & {{$\Delta$}} & {{F1 (\%)}} & {{$\Delta$}} \\
         \midrule
         {Ours (Full model)} & {\textbf{64.5}} & {-} & {\textbf{37.2}} & {-} \\
        %  \midrule
         \quad {- w/o HRE} & {59.4} & {5.1} & {30.5} & {6.7} \\
         \quad {- w/o BERT} & {61.3} & {3.2} & {33.4} & {3.8} \\
        %  \quad {- w/o Termination} & {61.9} & {2.6} & {34.7} & {2.5} \\
         \quad {- w/o Soft-switch} & {62.0} & {2.5} & {35.1} & {2.1} \\
         \bottomrule
    \end{tabular}}
    \caption{Ablation studies on two benchmark datasets.}
    \label{ablation}
\end{table}

\begin{table}
    \centering
    % \small
    \scalebox{0.88}{
    \begin{tabular}{lcccc}
         \toprule
         & \multicolumn{2}{c}{{SMD}} & \multicolumn{2}{c}{{MultiWOZ 2.1}}  \\
         \midrule
         \multirow{2}{*}{{{Model}}} & {{Original}} & {{Unseen}} & {{Original}} & {{Unseen}} \\
         & {{F1 (\%)}} & {{F1 (\%)}} & {{F1 (\%)}} & {{F1 (\%)}} \\
         \midrule
         {GLMP} & {60.7} & {55.3} & {32.4} & {23.9} \\
         {GraphDialog} & {61.1} & {55.7} & {34.1} & {25.4} \\
         {DF-Net} & {62.7} & {57.2} & {35.1} & {26.5} \\
        %  \midrule
         {Ours (Full)} & {\textbf{64.5}} & {\textbf{61.1}} & {\textbf{37.2}} & {\textbf{32.8}}  \\
         \bottomrule
    \end{tabular}}
    \caption{Generalization test results on two datasets.}
    \label{generalization}
\end{table}

\subsection{Generalization Capability}
We further investigate the generalization ability of our model under 
unseen settings. In the original dataset released by prior works, the 
entity overlap ratio between the train and test split is 78\% and 15.3\% 
for {MultiWOZ 2.1} and {SMD}, respectively. To simulate unseen scenario, we 
construct a new dataset split that reduces the entity overlap ratio to 
30\% for {MultiWOZ 2.1} and 2\% for {SMD} between the train and test split, 
which is a more challenging setting for all the models. More details of 
the construction process can be found in Appendix D. We re-run all the 
baselines with their released codes and our model on the new data split 
and report the results in Table \ref{generalization}. As we can see, the 
performance drops significantly for all systems on both datasets. However, our model 
degrades less compared to other systems, showing that it has better 
generalisation capability under unseen scenarios. This also verifies that neuro-symbolic
approach has the advantage of better generalisation ability which
has also been confirmed by many other studies \cite{andreas2016neural,rocktaschel2017end,minervini2020learning}.

\subsection{Human Evaluation}
Following prior work \cite{qin-etal-2020-dynamic}, we also conduct human 
evaluations for our framework and baselines from three aspects: 
\textit{Correctness}, \textit{Fluency}, and \textit{Humanlikeness}. Details 
about the scoring criterions can be found in Appendix H. We 
randomly select 300 different dialogue samples from the test set and ask 
human annotators to judge the quality of the responses and score them 
according to the three metrics ranging from 1 to 5. We train the annotators by showing them examples to help them understand the criteria 
and employ Fleiss’ kappa \cite{fleiss1971measuring} to measure the agreement across different annotators. The results are shown in Table \ref{human_study}. 
As we can see, our model outperforms all baselines across all the three metrics, consistent with our previous observations using automatic evaluations.

\begin{table}
    \centering
    % \small
    \scalebox{0.90}{
    \begin{tabular}{lccc}
        \toprule
        {{Model}} & {{Correct}} & {{Fluent}} & {{Humanlike}}  \\
        \midrule
        {GLMP} & {4.01} & {3.78} & {3.25}  \\
        {GraphDialog}  & {4.15} & {4.19} & {3.40}  \\
        {DF-Net} & {4.16} & {4.25} & {3.54}  \\
        {Ours (Full model)}  & {\textbf{4.41}} & {\textbf{4.28}} & {\textbf{3.59}}  \\
        \midrule
        {Human} & {4.83} & {4.65} & {4.57}  \\
        \midrule
        {Agreement} & {75\%} & {69\%} & {71\%} \\
        \bottomrule
    \end{tabular}}
    \caption{Human evaluation results.}
    \label{human_study}
\end{table}

\section{Conclusion}
In this paper, we propose an explicit and interpretable Neuro-Symbolic 
KB reasoning framework for task-oriented dialogue generation. The 
hypothesis generator employs a divide-and-conquer strategy to learn to 
generate hypotheses, and the reasoner employs a recursive strategy to 
learn to generate verification for the hypotheses. We evaluate our 
proposed framework on two public benchmark datasets including SMD and 
MultiWOZ 2.1. Extensive experimental results demonstrate the effectiveness of our 
proposed framework, as well being more interpretable.

\section{Ethical Considerations}\label{ethical}
For the human evaluation in this paper, we recruit several annotators on Amazon Mechanical Turk from English-speaking countries. We pay the annotators USD\$0.15 for each annotation task. Each task can be finished on average in 1 minute, which amounts to \$9.0 per hour that is above the US federal minimum wage (\$7.25). To ensure the quality of the human evaluation results, we perform quality control in a few ways. First, the annotators will be shown our scoring standards (Appendix H) before their tasks, and are asked to follow them. If the task is not done properly, either as determined by expert judgements (we recruit 3 native English speakers to validate the results of the Turkers' annotations) or there are obvious patterns such as constantly giving the same score for all tasks, we remove their annotations. We also compute agreement score to check for the consistency among the annotators.

\bibliography{anthology,custom}
\bibliographystyle{acl_natbib}

\appendix

% \section{Appendix}
\clearpage

\section{Details on Token Decoding in HRE}
Given the vector representations of the generated sub-hypotheses in hierarchical reasoning engine module, we utilize the similarity-based approach to decode the symbolic representations of those sub-hypotheses. Specifically, given a generated sub-hypotheses $[h_H, h_R, h_T]$, where $h_H$, $h_R$ and $h_T$ are the vector representations for the head entity, relation and tail entity correspondingly. To decode the symbolic representations for the head, relation and tail entities, we use:

\begin{equation}
	\mathop{\arg\min}_{\forall i} \ \ \| \phi(K_i)-h_H\|^{2}.
\end{equation}

\begin{equation}
	\mathop{\arg\min}_{\forall j} \ \ \| \phi(K_j)-h_R\|^{2}.
\end{equation}

\begin{equation}
	\mathop{\arg\min}_{\forall k} \ \ \| \phi(K_k)-h_T\|^{2}.
\end{equation}

\noindent where \textit{i}, \textit{j} and \textit{k} are the indices for the head entity, relation and tail entity in the vocabulary, $K_i$, $K_j$, $K_k$ denotes the $i$-th, $j$-th, $k$-th token of the KB, $\phi(K_i)$ denotes the embedding of the $i$-th token. Through this, we can decode the generated sub-hypotheses and obtain their explicit symbolic representations.

\section{Details on KB Distribution Calculation}
We extract the KB distribution $P_{kb,t}$ at timestep $t$ from the generated hypotheses and their corresponding belief scores as follows. For instance, if the generated hypothesis $[H, R, T]$ is an H-Hypothesis with a belief score $\alpha$, we extract the candidate token of the H-Hypothesis which is $H$ and then pair $H$ with the belief score $\alpha$, where $\alpha$ is viewed as the probability of the token $H$ to be selected as the output at timestep $t$. We conduct this for all the generated hypotheses and their corresponding belief scores from the HG and HRE modules. Finally, all the candidate tokens paired with their belief scores form the $P_{kb,t}$ at timestep $t$.

\section{Experimental Settings}
The dimensionality of the embedding and the decoder RNN hidden units are 
128 and embeddings are randomly initialized. The dropout ratio is 
selected from [0.1, 0.5]. We use Adam \cite{kingma2014adam} 
optimizer to optimize the parameters in our model and the learning rate 
is selected from [\textit{$1e^{-3}$},\textit{$1e^{-4}$}]. For the 
encoder, we fine-tune the BERT-base-uncased model from HuggingFace's 
library with an the embedding size of 768 with 12 layers and 12 heads.  
The maximum depth $D$ of the HRE module is selected from [1,5], the 
maximum number of candidates $K$ in CP module is selected from [1,10], 
and the temperature of Gumbel-Softmax is 0.1. All hyper-parameters are 
selected according to the validation set, and we repeat all the 
experiments 5 times with different random seeds and report the average 
results.

\section{Details on Unseen Setting}
We construct new dataset splits both on SMD and MultiWOZ 2.1 to simulate unseen scenarios for testing the generalization ability of all the models. Specifically, we construct the new dataset split as follows: We first extract all the KB entities that appeared in the dialogue responses and accumulate the percentage of samples for each KB entity. Second, we rank all the entities according to their percentage of samples in a decreasing order. Next, we split the KB entity set into train entities and test entities by accumulating the total percentages of samples. Finally, we iterate each sample in the dataset and assign it to train or test split by checking whether the entity in the response belong to the train entities or test entities. In this way, we obtain a new dataset split for both SMD and MultiWOZ 2.1, which has an entity overlap ratio of 2\% and 30\%, respectively, between train and test split (overlap ratio in the original SMD and MultiWOZ 2.1 are 15.3\% and 78\%, respectively).

The dataset statistics for the unseen splits are shown in Table \ref{dataset_statistics_3} and Table \ref{dataset_statistics_4}:

\begin{table}[!ht]
    \centering
    \begin{tabular}{c|c|c|c}
         \toprule[1pt]
         \textbf{Dataset} & \textbf{Train} & \textbf{Dev} & \textbf{Test}  \\
         \midrule
         SMD & 1850 & 311 & 870 \\
         \midrule
         MultiWOZ 2.1 & 1472 & 252 & 373 \\
         \bottomrule[1pt]
    \end{tabular}
    \caption{Statistics of Unseen Dataset for SMD and MultiWOZ 2.1.}
    \label{dataset_statistics_3}
\end{table}

\begin{table}[!ht]
    \centering
    \scalebox{0.8}{
    \begin{tabular}{cccc}
         \toprule[1pt]
         \multirow{2}{*}{\textbf{Dataset}} & Ent. Overlap & Ent. Overlap & \multirow{2}{*}{$\Delta$ $\downarrow$} \\
         & Standard & Unseen &  \\
         \midrule
         SMD & 15.3\% & 2\% & 13.3\% \\
         \midrule
         MultiWOZ 2.1 & 78\% & 30\% & 48\% \\
         \bottomrule[1pt]
    \end{tabular}}
    \caption{Entity Overlap Ratio Comparisons Between Unseen Split and Original Split for SMD and MultiWOZ 2.1. Entity Overlap Ratio = $\vert$Train Entities $\bigcap$ Test Entities$\vert$ / $\vert$Total Entities$\vert$.}
    \label{dataset_statistics_4}
\end{table}

\section{Additional Experiments}
We find that KB reasoning for most existing task-oriented dialogue datasets are quite simple, for the most part only requiring that only one or two hop reasoning over the KB in order to answer the user's request successfully. To further test our model and baseline models' multi-hop reasoning capability under complex reasoning scenarios, we develop a large-scale multi-domain synthetic dataset consisting dialogues requiring multi-hop reasoning over KBs. This is similar in spirit to bAbI dataset, and we hope that this dataset will continue to be used with other dialogue benchmarks in future studies. We will release this dataset upon publication. Next, we describe how we construct the dataset in details and show the experimental results performed on it.

\subsection{Dataset Construction}
\begin{figure}[!ht]
    \centering
    \includegraphics[width=3.0in]{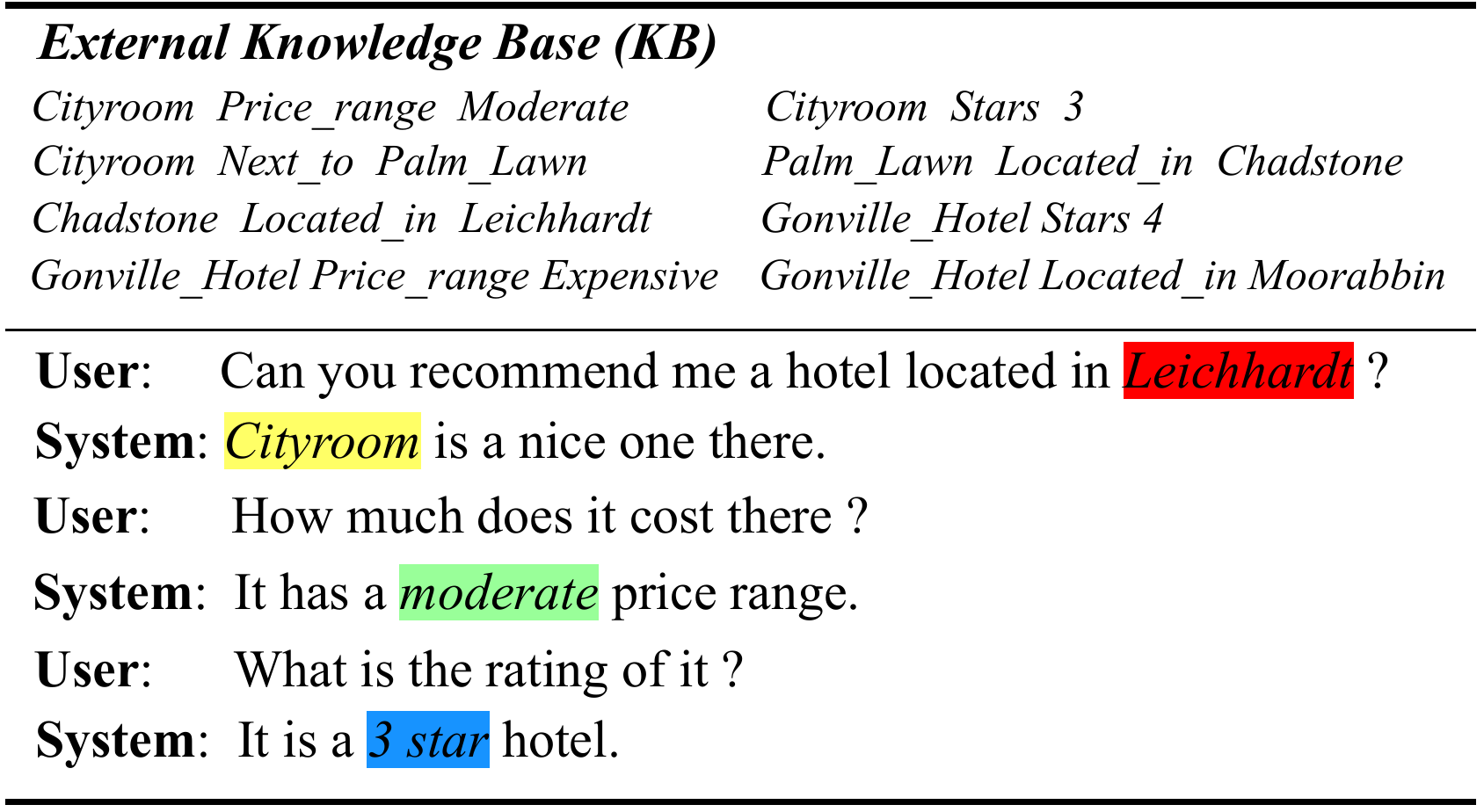}
    \caption{An example dialogue from the hotel domain of the synthetic dataset. The first turn of the dialogue requires a 3-hop reasoning over the KB to get the correct entity Cityroom given the location information Leihhardt. The second and third turn of the dialogue require single-hop reasoning over KB to get the correct entity.}
    \label{example_dialogue_5}
\end{figure}

As is shown in Figure \ref{example_dialogue_5}, each sample in the dataset consists of several rounds of dialogues. We generate the questions and answers of the dialogues by randomly sample template utterances with placeholders (e.g., \textit{@movie}, \textit{@director}, \textit{@location}) indicating the types of KB entities to be instantiated to form the complete utterances. To simulate a natural conversation between user and system under different scenarios (i.e., restaurant booking, hotel reservation, movie booking), we designed 18 different types of question-answer templates. For example, \textit{movie to director} denotes that the user requests the director given the movie name, \textit{location to theatre} denotes the user requires theatre information given the location. For each conversation, we randomly select several different types of question-answer templates sequentially to form the skeleton of the whole dialogue. To ensure the coherent of the dialogue flow, we provide the guided next types for each question-answer template. For instance, if the current sampled question-answer type is \textit{location to restaurant}, the guided next types will be randomly sampled from \textit{restaurant to price}, \textit{restaurant to cuisine} etc. Thus, we can ensure the generated dialogue turns more coherent in terms of semantics to simulate a real conversation as much as possible.

For each conversation, we generate 3 or 4 rounds of dialogues following the existing work such as \textit{SMD} and \textit{MultiWOZ 2.1}. At each round of the dialogue, we randomly select a question-answer template and instantiate the placeholders in the template with the corresponding types of KB entities. If there are multiple entities in the KB satisfy the types indicated by the placeholders, we randomly sample one to implement the template. In this way, we can increase the diversity of the generated data. For instance, if the question template is \textit{Is there any restaurant located in @district?}, the possible sets of entities in the KB for the placeholder \textit{@district} might include multiple location entities in the KB such as \textit{vermont}, \textit{blackburn} etc. We randomly sample one of them to replace the placeholder and generate a final sentence. If we sample \textit{vermont}, the implemented sentence will be \textit{Is there any restaurant located in the vermont?}.

To make the generated dialogue utterances more natural as human conversations, we further randomly replace the KB entities in the sentence with pronouns such as \textit{it}, \textit{they} etc, provided that the entities have been mentioned in previous dialogue turns. Thus, it requires the model to overcome the co-reference resolution to arrive at the correct answer which increases the difficulty. For example, \textit{Who is the director of the movie mission impossible?} will be rephrased as \textit{Who is the director of it?} if the movie name \textit{mission impossible} has been mentioned in the dialogue history.

For movie domain, we employ the KB used in the well-known \textit{WikiMovie} dataset. For hotel and restaurant domain, we use the KB provided in the \textit{MultiWOZ 2.1} dataset. For each employed KB, we further extend it by adding information such as hierarchies of locations to enrich the KB in order to make it suitable for testing multi-hop reasoning capability. For example, if the KB contains a hotel entity \textit{love lodge}, we add different levels of location information to support multi-hop KB reasoning. For instance, we add location information such as \textit{love\_lodge next\_to lincoln\_park}, \textit{lincoln\_park is\_within waverley\_district}, \textit{waverley\_district located\_in grattan\_county}. Thus, if the user asked about the hotel located in \textit{grattan\_county}, it requires the model to conduct multi-hop reasoning over the KB to know that \textit{love\_lodge}is located in \textit{grattan\_county}. Through this, we make our synthetic dataset suitable for multi-hop reasoning tasks over KB under task-oriented dialogue scenarios. The location information we utilized in the synthetic dataset are obtained from the Wikipedia and the official website of famous cities around the world.

\subsection{Dataset Statistics}
The detailed statistics of the synthetic dataset are shown in Table \ref{dataset_statistics_1} and Table \ref{dataset_statistics_2}:

\begin{table}[!ht]
    \centering
    \begin{tabular}{cccc}
         \toprule[1pt]
         \textbf{Domain} & \textbf{Train} & \textbf{Dev} & \textbf{Test}  \\
         \midrule
         Movie & 7219 & 1645 & 1667 \\
         \midrule
         Hotel & 7115 & 1631 & 1639  \\
         \midrule
         Restaurant & 7131 & 1672 & 1684 \\
         \midrule
         Total & 21465 & 4948 & 4990 \\
         \bottomrule[1pt]
    \end{tabular}
    \caption{Statistics of synthetic dataset. Numbers in the table are the number of instances for each category.}
    \label{dataset_statistics_1}
\end{table}

\begin{table*}[!t]
    \centering
    \begin{tabular}{cccccccccc}
         \toprule[1pt]
          & \multicolumn{3}{c}{\textbf{{Movie}}} &  \multicolumn{3}{c}{\textbf{{Hotel}}} &  \multicolumn{3}{c}{\textbf{{Restaurant}}}  \\
          \midrule
          & \textbf{Train} & \textbf{Dev} & \textbf{Test} & \textbf{Train} & \textbf{Dev} & \textbf{Test} & \textbf{Train} & \textbf{Dev} & \textbf{Test}  \\
         \midrule
         \textbf{Hop=1} & 27238 & 5985 & 5998 & 14321 & 3472 & 3482 & 14386 & 2107 & 2117 \\
         \midrule
         \textbf{Hop=2} & 6401 & 1472 & 1507 & 3351 & 594 & 614  & 4527 & 564 & 609 \\
         \midrule
         \textbf{Hop>=3} & 5359 & 1508 & 1530 & 3328 & 514 & 524  & 4545 & 562 & 593 \\
         \midrule
         \textbf{Total} & 38998 & 8965 & 9035 & 21000 & 4580 & 4620 & 23458 & 3233 & 3319 \\
         \bottomrule[1pt]
    \end{tabular}
    \caption{Detailed statistics of the synthetic dataset with respect to the number of hops needed by KB reasoning. Numbers in the table are the number of dialogue turns for each category. Hop=$k$ denotes that the KB reasoning path length for the entity in the dialogue response is $k$.}
    \label{dataset_statistics_2}
\end{table*}

\subsection{Experimental Results}
\noindent \textbf{Evaluation Metrics.}
We use the same metrics as on \textit{SMD} and \textit{MultiWOZ 2.1} dataset includes \textit{BLEU} and \textit{Entity F1} for performance evaluation.

\noindent \textbf{Results.} The results on the three domains are shown in Table \ref{movie_result}, \ref{hotel_result}, \ref{restaurant_result}. For each domain, we evaluate the model performance on different subsets of the test data, i.e., 1-hop, 2-hop and >=3-hop. Specifically, we group the test data into three different subsets according to the KB reasoning length for obtaining the ground-truth entity. For instance, 2-hop denotes that the KB entity mentioned in the response needs 2-hop reasoning over the KB. As we can see from the tables, our proposed model consistently outperforms all the baselines by a large margin across all the domains and KB reasoning lengths. We also observe that all the models' performance decrease monotonously as the KB reasoning path length increases, suggesting that KB reasoning with longer range is challenging for all the tested models. However, our framework has less performance degradation compared to all the baselines, and the performance gap between our framework and the baselines has become larger when the length of KB reasoning increases, which demonstrates that our framework has better generalization ability especially under longer KB reasoning paths compared to those baselines.

% (from 2-Hop to 4-Hop)

\begin{table*}[!t]
    \centering
    \begin{tabular}{rccccccccc}
         \toprule[1pt]
          & \multicolumn{8}{c}{{\textbf{Movie Domain}}} \\
          \midrule
          & \multicolumn{2}{c}{\textbf{1-Hop}} & \multicolumn{2}{c}{\textbf{2-Hop}} & \multicolumn{2}{c}{\textbf{Hop>=3}} & \multicolumn{2}{c}{\textbf{All}}  \\
         \midrule
         & BLEU & F1 & BLEU & F1 & BLEU & F1 & BLEU & F1  \\
         \midrule
         Mem2Seq & 25.6 & 68.9 & 21.8 & 60.8 & 19.5 & 49.2 & 23.9 & 62.0  \\
         GLMP & 30.1 & 77.2 & 28.7 & 72.9 & 27.1 &  61.5 & 28.3 & 73.2   \\
         GraphDialog & 29.2 & 76.6 & 25.6 & 69.1 & 24.7 &  60.6 & 27.2 & 71.6  \\
         DF-Net & 30.6 & 77.4 & 29.5 & 71.6 & 28.9 & 62.1 & 30.3 &  73.5  \\
         \midrule
         \textbf{Ours (Full model)} & \textbf{33.2} & \textbf{82.6} & \textbf{31.3} & \textbf{80.4} & \textbf{30.7} & \textbf{74.9} & \textbf{32.7} &  \textbf{80.6}  \\
         \bottomrule[1pt]
    \end{tabular}
    \caption{Experimental results on the movie domain of the synthetic dataset.}
    \label{movie_result}
\end{table*}

\begin{table*}[!t]
    \centering
    \begin{tabular}{rccccccccc}
         \toprule[1pt]
          & \multicolumn{8}{c}{{\textbf{Hotel Domain}}} \\
          \midrule
          & \multicolumn{2}{c}{\textbf{1-Hop}} & \multicolumn{2}{c}{\textbf{2-Hop}} & \multicolumn{2}{c}{\textbf{Hop>=3}} & \multicolumn{2}{c}{\textbf{All}}  \\
         \midrule
         & BLEU & F1 & BLEU & F1 & BLEU & F1 & BLEU & F1  \\
         \midrule
         Mem2Seq & 14.4 & 79.8 & 13.1 & 71.2 & 11.4 & 68.6 & 13.2 & 75.4  \\
         GLMP & 21.3 & 85.5 & 19.8 & 79.4 & 18.9 & 76.2  & 21.0 &  82.9  \\
         GraphDialog & 20.6 & 83.8 & 19.1 & 78.8 & 18.8 & 75.9  & 19.3 & 81.0  \\
         DF-Net & 22.1 & 86.7 & 19.9 & 80.2 & 19.5 & 76.8 & 21.5 &  83.2  \\
         \midrule
         \textbf{Ours (Full model)} & \textbf{23.3} & \textbf{92.4} & \textbf{21.3} & \textbf{89.6} & \textbf{20.7} & \textbf{87.8} & \textbf{22.1} &  \textbf{91.6}  \\
         \bottomrule[1pt]
    \end{tabular}
    \caption{Experimental results on the hotel domain of the synthetic dataset.}
    \label{hotel_result}
\end{table*}

\begin{table*}[!t]
    \centering
    \begin{tabular}{rccccccccc}
         \toprule[1pt]
          & \multicolumn{8}{c}{{\textbf{Restaurant Domain}}} \\
          \midrule
          & \multicolumn{2}{c}{\textbf{1-Hop}} & \multicolumn{2}{c}{\textbf{2-Hop}} & \multicolumn{2}{c}{\textbf{Hop>=3}} & \multicolumn{2}{c}{\textbf{All}}  \\
         \midrule
         & BLEU & F1 & BLEU & F1 & BLEU & F1 & BLEU & F1  \\
         \midrule
         Mem2Seq  & 19.0 & 79.8 & 17.3 & 69.4 & 12.4 & 66.3 & 17.0 & 73.7  \\
         GLMP & 22.0 & 90.4 & 19.1 & 83.7 & 18.4 & 80.4  & 20.9 &  86.1  \\
         GraphDialog & 23.2 & 89.9 & 21.2 & 82.1 & 20.6 & 79.8  & 21.4 & 85.0  \\
         DF-Net & 24.5 & 91.5 & 23.0 & 84.2 & 21.1 & 81.0 & 23.3 & 87.3   \\
         \midrule
         \textbf{Ours (Full model)} & \textbf{26.8} & \textbf{96.7} & \textbf{24.4} & \textbf{93.1} & \textbf{22.7} & \textbf{92.2} & \textbf{25.1} & \textbf{94.2}   \\
         \bottomrule[1pt]
    \end{tabular}
    \caption{Experimental results on the restaurant domain of the synthetic dataset.}
    \label{restaurant_result}
\end{table*}

\subsection{Example Outputs}
\begin{figure*}[!ht]
    \centering
    \includegraphics[width=6.5in]{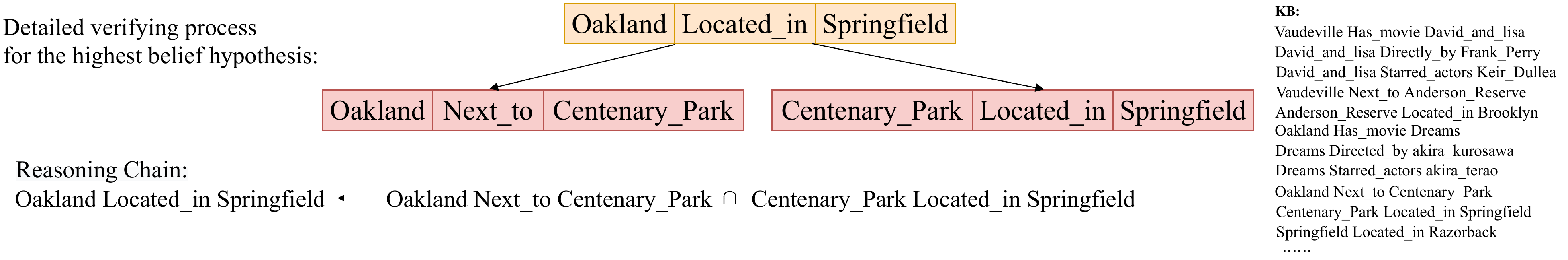}
    \caption{Proof tree generated by HRE module for the highest score hypothesis \textit{[Oakland, Located\_in, Springfield]} in Table \ref{case_study_1}. The red parts are the predicted bridge entities and the blue parts are the predicted relations for the sub-hypotheses via neural networks. In this case, the model performs 2-hop reasoning (the two leaf node triples) to find the correct KB entity for generating the response.}
    \label{example_dialogue_1}
\end{figure*}

We show the generated hypotheses and the proof trees in our framework as shown in Table \ref{case_study_1} and Figure \ref{example_dialogue_1}. As we can see, our model can successfully obtain the correct entities from the KB. Moreover, our framework can formulate sensible hypotheses and generate reasonable proof procedures which can help us gain some insights about the inner workings of our model.

\begin{table*}[!t]
    \centering
    \scalebox{0.9}{
    \begin{tabular}{ccccc}
         \toprule[1pt]
         \textbf{Structure} & \textbf{State} & \textbf{Top-5 Candidate} & \textbf{Generated} & \textbf{Belief} \\
         \textbf{Type} & \textbf{Tokens} & \textbf{Tokens} & \textbf{Hypotheses} & \textbf{Scores} \\
         \midrule
         \multirow{5}{*}{\textit{H-Hypothesis}} &  & ``\textit{Shipping\_News}'' & ``\textit{{[Shipping\_News, located\_in, Springfield]}}'' & {0.15}  \\
          & ``\textit{located\_in} ($k$=0)''  & ``\textit{Vaudeville}'' & ``\textit{[Vaudeville, located\_in, Springfield]}'' & 0.17 \\
          &   & ``\textit{Brown\_Eyes}'' & ``\textit{[Brown\_Eyes, located\_in, Springfield]}'' & 0.13 \\
          & ``\textit{Springfield} ($k$=1)'' & ``\textit{Oakland}'' & ``\textit{[Oakland, located\_in, Springfield]}'' & 1.00 \\
          &  & ``\textit{Coburg}'' & ``\textit{[Coburg, located\_in, Springfield]}'' & 0.04 \\
         \bottomrule[1pt]
    \end{tabular}}
    \caption{Example outputs on the movie domain of synthetic dataset. Dialogue history: ``\textit{I'm looking for a theatre in the Springfield district.}''. Generated response: ``\textit{Sure I have found a Oakland for you.}''. Detailed model working process when generating Oakland in the response is shown above.}
    \label{case_study_1}
\end{table*}

\section{Error Analysis}
We conduct error analysis on both \textit{SMD} and \textit{MultiWOZ 2.1} to provide insights in our 
framework for future improvements. We randomly sample 100 dialogues from each test set and analysis both the generated responses and the inner procedures.
The errors have four 
major categories: 1) structure errors, 2) query states errors, 3) candidates errors, 4) 
belief score errors. For example, given dialogue history ``\textit{Where is a nearby parking\_garage?}'', the generated response is ``\textit{5671\_barringer\_street is 1\_mile away.}'' and the ground-truth is ``\textit{The nearest one is palo\_alto\_garage, it's just 1\_mile away.}''. The generated hypotheses by HG module at timestep 0 are ``[\textit{5671\_barringer\_street}, \textit{distance}, \textit{1\_mile}], [\textit{whole\_foods}, \textit{distance}, \textit{1\_mile}], [\textit{home}, \textit{distance}, \textit{1\_mile}], [\textit{valero}, \textit{distance}, \textit{1\_mile}], [\textit{pizza\_chicago}, \textit{distance}, \textit{1\_mile}]''. The predicted structure type is a H-Hypothesis with a relation ``\textit{distance}'' and tail entity ``\textit{1\_mile}''. However, a reasonable hypothesis should be an H-Hypothesis such as ``[\textit{palo\_alto\_garage}, \textit{poi\_type}, \textit{parking\_garage}]'' with ``\textit{poi\_type}'' (place of interest) and ``\textit{parking\_garage}'' as relation and tail. This illustrates the challenge of the task, and we plan to further improve our framework include SP, QSP, CP and HRE in future work.

\subsection{Structure Prediction Errors}
Given the dialogue history ``\textit{Find me a nearby shopping mall.}'', the generated response is ``\textit{midtown\_shopping\_center is 3\_miles away.}'' and the ground-truth response is ``\textit{The closest shopping\_center is midtown\_shopping\_center 3\_miles away.}''. The generated hypotheses by HG module at decoding timestep 0 are ``\trp{home, address, 5671\_barringer\_street}, \trp{home, address, 481\_amaranta\_ave}, \trp{home, address, midtown\_shopping\_center}, \trp{home, address, 5\_miles}, \trp{home, address, 1\_miles}''. As we can see, the predicted structure type is an T-Hypothesis with an head entity ``\textit{home}'' and relation ``\textit{address}''. However, a reasonable structure type should be an H-Hypothesis such as ``\trp{midtown\_shopping\_center, poi\_type, shopping\_center}'' which the model has mispredicted to a T-Hypothesis.

\subsection{Query States Prediction Errors}
Given the dialogue history ``\textit{What time is dinner tonight?}'', the generated response is ``\textit{Dinner is at 11pm.}'' and the ground-truth response is ``\textit{Your dinner is 5pm tonight.}''. The generated hypotheses in HG module for decoding timestep 3 are ``\trp{dinner, traffic\_info, 11pm}, \trp{dinner, traffic\_info, 7pm}, \trp{dinner, traffic\_info, 3pm}, \trp{dinner, traffic\_info, Monday}, \trp{dinner, traffic\_info, Tuesday}''. As we can see, the SP and QSP modules have successfully predicted a sensible T-Hypothesis with ``\textit{dinner}'' as head entity. However, the predicted relation ``\textit{traffic\_info}'' is not reasonable according to context where an expected relation would be relations such as ``time''.

\subsection{Candidates Prediction Errors}
Given the dialogue history ``\textit{What time is my doctor\_appointment?}'', the generated response is ``\textit{Your doctor\_appointment is on Thursday.}'' and the ground-truth response is ``\textit{Your doctor\_appointment is scheduled for Friday at 11am.}''. The generated hypotheses by HG for decoding timestep 4 are ``\trp{doctor\_appointment, date, Thursday}, \trp{doctor\_appointment, date, Monday}, \trp{doctor\_appointment, date, 7pm}, \trp{doctor\_appointment, date, Tuesday}, \trp{doctor\_appointment, date, 2pm}''. As we can see, our framework has correctly predicted a T-Hypothesis and state tokens. However, the predicted top-5 candidate tokens don't include the ground-truth date which is \textit{Friday}, which might affect the final response generation.

\subsection{Belief Score Prediction Errors}
Given the dialogue history ``\textit{Where is home?}'', the generated response is ``\textit{Home is at 819\_alma\_st.}'' and the ground-truth is ``\textit{Home is where is your heart, at 56\_cadwell\_street.}'' The generated hypotheses for decoding timestep 3 are ``\trp{home, address, 819\_alma\_st}, \trp{home, address, 56\_cadwell\_street}, \trp{home, address, 6\_miles}, \trp{home, address, 611\_ames\_ave}, \trp{home, address, 3\_miles}''. As we can see, our framework has predicted a sensible T-Hypothesis with ``\textit{home}'' as head entity and ``\textit{address}'' as relation. Also, the CP module has predicted top-5 candidate tail entities which include the ground-truth \textit{56\_cadwell\_street}. But the HRE module ranked ``\trp{home, address, 819\_alma\_st}'' highest with a score of 0.78 while the ground-truth one ``\trp{home, address, 56\_cadwell\_street}'' is only ranked the second highest with a score of 0.41, which indicates that there is still room for improvements for the HRE module. We are interested in continually improving our framework include all the modules in future work.

\section{Discussions}
\subsection{Why not use search-based techniques for generating reasoning chains?}
This is an alternative approach to our learning-based method. However, search-based approach
cannot be jointly learnt end-to-end with other modules in our framework, and thus may face
error propagation and credit assignment issues like in the traditional pipeline-based task-oriented
dialogue approaches. In this work, we want to explore the possibility of  learning end-to-end
the logical reasoning chain directly from the dialogues. Also, the time complexity of search-based approach is approximately $O(n^{k})$, where $n$ is the
average degree of nodes in the external knowledge base, $k$ is the number of reasoning hops. In other words, the 
time complexity tends to have polynomial growth (when $k > 1$); and it's worse when the reasoning complexity ($k$) increases (exponential).
In contrast, when the number of KB nodes increases it only 
impacts the size of the input embedding layer in our framework  (Equation \ref{embed_layer}), and the efficiency can be further improved by leveraging modern accelerating hardware such as GPU (which search-based approaches cannot).

\subsection{Why sample with Gumbel-Softmax instead of directly applying argmax in Hypothesis Generator and Hierarchical Reasoning Engine modules?}
Argmax function is non-differentiable which hinders our aim of end-to-end differentiability of the whole system. We tried utilizing REINFORCE (reward is obtained by comparing predicted entities with ground-truth entities) to mitigate this issue. However, we find that the results of using argmax+REINFORCE is worse than using Gumbel-Softmax. By checking the sampled tokens from Gumbel-Softmax, we find that it can generate reasonable tokens (Figure \ref{example_dialogue_2}
%JHL5: need to reference properly
in the main paper, state tokens etc.), since we have set the temperature parameter of Gumbel-Softmax to 0.1 which is a close approximation to argmax.

\subsection{Why not expand the KB using KB completion methods and then use semantic parsing to query KB?}
In this work, we are interested in developing an end-to-end trainable framework with explainable KB reasoning. Semantic parsing is one possible alternative. However, when adapting to our own dataset, it requires further annotations for fine-tuning which is costly and time-consuming, and might be not feasible for large-scale datasets. Also, it might induce the error propagation issue since the different modules (KB completion, semantic parsing, dialogue encoding and response generation etc.) are not jointly learnt.

\subsection{KB scale.}
The average nodes of KB for each sample in the training data is 63.5 for \textit{SMD} and 57.6 for \textit{MultiWOZ}.
The average number of relations is 5.5 for \textit{SMD} and 9.4 for \textit{MultiWOZ}.

\section{Human Evaluation Details}
The \textit{Fluency} of the predicated responses is evaluated according to the following standards:
\begin{itemize}
    \item 5: The predicted responses contain no grammar errors or repetitions at all.
    \item 4: Only one grammar error or repetition appeared in the generated responses.
    \item 3: One grammar error one repetition, or two grammar errors, or two repetitions are observed in the responses.
    \item 2: One grammar error two repetitions, or one repetition two grammar errors, or three grammar errors, or three repetitions appeared in the generated responses.
    \item 1: More than three inappropriate language usages with regard to grammar errors or repetitions are observed in the responses.
\end{itemize}

The \textit{Correctness} is measured as follows:
\begin{itemize}
    \item 5: Provide the correct entities.
    \item 4: Minor mistakes in the provided entities.
    \item 3: Noticeable errors in the provided entities but acceptable.
    \item 2: Poor in the provided entities.
    \item 1: Wrong in the provided entities.
\end{itemize}

The \textit{Humanlikeness} is measured as:
\begin{itemize}
    \item 5: 100\% sure that the sentences are generated by a human, not by system.
    \item 4: 80\% chance that the sentences are generated by a human.
    \item 3: Cannot tell whether the sentences is generated by a human or system, 50\% for human and 50\% for system.
    \item 2: 20\% chance that the sentences are generated by a human.
    \item 1: Totally impossible that the sentences are generated by a human.
\end{itemize}

\end{document}

\appendix

\section{Details on Token Decoding in HRE}
Given the vector representations of the generated sub-hypotheses in hierarchical reasoning engine module, we utilize the similarity-based approach to decode the symbolic representations of those sub-hypotheses. Specifically, given a generated sub-hypotheses $[h_H, h_R, h_T]$, where $h_H$, $h_R$ and $h_T$ are the vector representations for the head entity, relation and tail entity correspondingly. To decode the symbolic representations for the head, relation and tail entities, we use:

\begin{equation}
	\mathop{\arg\min}_{\forall i} \ \ \| \phi(K_i)-h_H\|^{2}.
\end{equation}

\begin{equation}
	\mathop{\arg\min}_{\forall j} \ \ \| \phi(K_j)-h_R\|^{2}.
\end{equation}

\begin{equation}
	\mathop{\arg\min}_{\forall k} \ \ \| \phi(K_k)-h_T\|^{2}.
\end{equation}

\noindent where \textit{i}, \textit{j} and \textit{k} are the indices for the head entity, relation and tail entity in the vocabulary, $K_i$, $K_j$, $K_k$ denotes the $i$-th, $j$-th, $k$-th token of the KB, $\phi(K_i)$ denotes the embedding of the $i$-th token. Through this, we can decode the generated sub-hypotheses and obtain their explicit symbolic representations.

\section{Details on KB Distribution Calculation}
We extract the KB distribution $P_{kb,t}$ at timestep $t$ from the generated hypotheses and their corresponding belief scores as follows. For instance, if the generated hypothesis $[H, R, T]$ is an H-Hypothesis with a belief score $\alpha$, we extract the candidate token of the H-Hypothesis which is $H$ and then pair $H$ with the belief score $\alpha$, where $\alpha$ is viewed as the probability of the token $H$ to be selected as the output at timestep $t$. We conduct this for all the generated hypotheses and their corresponding belief scores from the HG and HRE modules. Finally, all the candidate tokens paired with their belief scores form the $P_{kb,t}$ at timestep $t$.

\section{Details on Unseen Setting}
We construct new dataset splits both on SMD and MultiWOZ 2.1 to simulate unseen scenarios for testing the generalization ability of all the models. Specifically, we construct the new dataset split as follows: We first extract all the KB entities that appeared in the dialogue responses and accumulate the percentage of samples for each KB entity. Second, we rank all the entities according to their percentage of samples in a decreasing order. Next, we split the KB entity set into train entities and test entities by accumulating the total percentages of samples. Finally, we iterate each sample in the dataset and assign it to train or test split by checking whether the entity in the response belong to the train entities or test entities. In this way, we obtain a new dataset split for both SMD and MultiWOZ 2.1, which has an entity overlap ratio of 2\% and 30\%, respectively, between train and test split (overlap ratio in the original SMD and MultiWOZ 2.1 are 15.3\% and 78\%, respectively).

The dataset statistics for the unseen splits are shown in Table \ref{dataset_statistics_3} and Table \ref{dataset_statistics_4}:

\begin{table}[!ht]
    \centering
    \begin{tabular}{c|c|c|c}
         \toprule[1pt]
         \textbf{Dataset} & \textbf{Train} & \textbf{Dev} & \textbf{Test}  \\
         \midrule
         SMD & 1850 & 311 & 870 \\
         \midrule
         MultiWOZ 2.1 & 1472 & 252 & 373 \\
         \bottomrule[1pt]
    \end{tabular}
    \caption{Statistics of Unseen Dataset for SMD and MultiWOZ 2.1.}
    \label{dataset_statistics_3}
\end{table}

\begin{table}[!ht]
    \centering
    \begin{tabular}{cccc}
         \toprule[1pt]
         \multirow{2}{*}{\textbf{Dataset}} & Ent. Overlap & Ent. Overlap & \multirow{2}{*}{$\Delta$ $\downarrow$} \\
         & Original & Unseen &  \\
         \midrule
         SMD & 15.3\% & 2\% & 13.3\% \\
         \midrule
         MultiWOZ 2.1 & 78\% & 30\% & 48\% \\
         \bottomrule[1pt]
    \end{tabular}
    \caption{Entity Overlap Ratio Comparisons Between Unseen Split and Original Split for SMD and MultiWOZ 2.1. Entity Overlap Ratio = $\vert$Train Entities $\bigcap$ Test Entities$\vert$ / $\vert$Total Entities$\vert$.}
    \label{dataset_statistics_4}
\end{table}

\section{Additional Experiments}
We further conduct additional experiments to test our model's multi-hop reasoning ability on a large-scale multi-domain synthetic dataset. Next, we describe how we build the synthetic dataset in details and show the experimental results on the dataset.

\subsection{Dataset Construction}

%JHL2: typo of leihhardt vs. leichhardt

As is shown in Figure \ref{example_dialogue_5}, each sample is consisted of multiple rounds of dialogues. We generate the questions and answers by randomly sample template utterances with placeholders (e.g., @movie, @director, @location) indicating the types of KB entities to be instantiated to form the complete utterances. To simulate a natural conversation between user and system under different scenarios, we've designed 18 different types of question-answer templates. For instance, ``movie to director'' denotes that the user requires the director given the movie name, ``location to theatre'' denotes the user requires theatre information given the location. For each conversation, we randomly select several different types of question-answer templates sequentially to form the skeleton of the whole dialogue. To ensure the coherent of the dialogue flow, we provide the guided next types for each question-answer template. For instance, if the current sampled question-answer type is ``location to restaurant'', the guided next types will be randomly sampled from ``restaurant to price'', ``restaurant to cuisine '' etc. Thus, we can ensure the generated dialogue turns more coherent in terms of semantics to simulate a real conversation as much as possible.

% Overall, we've included 15 different types of KB entities. The detailed types and its corresponding placeholders used in the template are shown in Table 1.
% Detailed question-answer templates we used can be found in Table 2. 

For each conversation, we generate 3 or 4 rounds of dialogues following the existing work such as SMD and MultiWOZ 2.1. At each round of the dialogue, we randomly select a question-answer template and instantiate the placeholders in the template with the corresponding types of KB entities. If there are multiple entities in the KB satisfy the types indicated by the placeholders, we randomly sample one to implement the template. In this way, we can increase the diversity of the generated data. For instance, if the question template is ``Is there any restaurant located in @district?'', the possible sets of entities in the KB for the placeholder ``@district'' might include multiple location entities in the KB such as ``vermont'', ``blackburn'' etc. We randomly sample one of them to replace the placeholder and generate a final sentence. If we sample ``vermont'', the implemented sentence will be ``Is there any restaurant located in the vermont?''.

To make the generated dialogue utterances more natural as human conversations, we further randomly replace the KB entities in the sentence with pronouns such as ``it'', ``they'' etc, provided that the entities have been mentioned in previous dialogue turns. Thus, it requires the model to overcome the co-reference resolution to arrive at the correct answer which increases the difficulty. For example, ``Who is the director of the movie mission impossible?'' will be rephrased as ``Who is the director of it?'' if the movie name ``mission impossible'' has been mentioned in the dialogue history.

For movie domain, we employ the KB used in the well-known WikiMovie dataset. The reason that we don't use the utterances in WikiMovie dataset is that the utterances are not originally designed for testing multi-hop reasoning ability. For hotel and restaurant domain, we use the KB provided in the MultiWOZ 2.1 dataset. For each employed KB, we further extend it by adding information such as hierarchies of locations to enrich the KB in order to make it suitable for testing multi-hop reasoning ability. For example, if the KB contains an hotel entity ``love lodge'', we add different levels of location information to support multi-hop KB reasoning. For instance, we add location information such as ``love\_lodge next\_to lincoln\_park'', ``lincoln\_park is\_within waverley\_district'', ``waverley\_district located\_in grattan\_county''. Thus, if the user asked about the hotel located in grattan\_county, it requires the model to conduct multi-hop reasoning over the KB to know that love\_lodge is located in grattan\_county. Through this, we make our synthetic dataset suitable for multi-hop reasoning tasks over KB under dialogue scenarios. The location information we utilized in the synthetic dataset are obtained from the Wikipedia and the official website of famous cities around the world.

\begin{figure}[!ht]
    \centering
    \includegraphics[width=3.0in]{example_dialogue_synthetic_2.pdf}
    \caption{An example dialogue from the hotel domain of the synthetic dataset. The first turn of the dialogue requires a 3-hop reasoning over the KB to get the correct entity Cityroom given the location information Leihhardt. The second and third turn of the dialogue require single-hop reasoning over KB to get the correct entity.}
    \label{example_dialogue_5}
\end{figure}

\subsection{Dataset Statistics}
The detailed statistics of the synthetic dataset are shown in Table \ref{dataset_statistics_1} and Table \ref{dataset_statistics_2}:

\begin{table}[!ht]
    \centering
    \begin{tabular}{cccc}
         \toprule[1pt]
         \textbf{Domain} & \textbf{Train} & \textbf{Dev} & \textbf{Test}  \\
         \midrule
         Movie & 7219 & 1645 & 1667 \\
         \midrule
         Hotel & 7115 & 1631 & 1639  \\
         \midrule
         Restaurant & 7131 & 1672 & 1684 \\
         \midrule
         Total & 21465 & 4948 & 4990 \\
         \bottomrule[1pt]
    \end{tabular}
    \caption{Statistics of synthetic dataset. Numbers in the table are the number of instances for each category.}
    \label{dataset_statistics_1}
\end{table}

\begin{table*}[!t]
    \centering
    \begin{tabular}{cccccccccc}
         \toprule[1pt]
          & \multicolumn{3}{c}{\textbf{{Movie}}} &  \multicolumn{3}{c}{\textbf{{Hotel}}} &  \multicolumn{3}{c}{\textbf{{Restaurant}}}  \\
          \midrule
          & \textbf{Train} & \textbf{Dev} & \textbf{Test} & \textbf{Train} & \textbf{Dev} & \textbf{Test} & \textbf{Train} & \textbf{Dev} & \textbf{Test}  \\
         \midrule
         \textbf{1-Hop} & 27238 & 5985 & 5998 & 14321 & 3472 & 3482 & 14386 & 2107 & 2117 \\
         \midrule
         \textbf{2-Hop} & 6401 & 1472 & 1507 & 3351 & 594 & 614  & 4527 & 564 & 609 \\
         \midrule
         \textbf{3-Hop} & 5359 & 1508 & 1530 & 3328 & 514 & 524  & 4545 & 562 & 593 \\
         \midrule
         \textbf{Total} & 38998 & 8965 & 9035 & 21000 & 4580 & 4620 & 23458 & 3233 & 3319 \\
         \bottomrule[1pt]
    \end{tabular}
    \caption{Detailed statistics of the synthetic dataset with respect to the number of hops needed by KB reasoning. Numbers in the table are the number of dialogue turns for each category. $k$-Hop denotes that the KB reasoning path length for the entity in the dialogue response is $k$.}
    \label{dataset_statistics_2}
\end{table*}

\subsection{Experimental Results}
\noindent \textbf{Evaluation Metrics.}
We use the same metrics as on SMD and MultiWOZ 2.1 dataset includes \textit{BLEU} and \textit{Entity F1} for performance evaluation.

\noindent \textbf{Results.} The results on the three domains are shown in Table \ref{movie_result}, \ref{hotel_result}, \ref{restaurant_result}. For each domain, we evaluate the model performance on different subsets of the test data, i.e., 1-hop, 2-hop and 3-hop. Specifically, we group the test data into three different subsets according to the KB reasoning length for obtaining the ground-truth entity. For instance, 2-hop denotes that the KB entity mentioned in the response needs 2-hop reasoning over the KB. As we can see from the tables, our proposed model consistently outperforms all the baselines by a large margin across all the domains and KB reasoning lengths. We also observe that all the models' performance decrease monotonously as the KB reasoning path length increases (from 1-Hop to 3-Hop), suggesting that KB reasoning with longer range is challenging for all the tested models. However, our framework has less performance degradation compared to all the baselines, and the performance gap between our framework and the baselines has become larger when the length of KB reasoning increases, which demonstrates that our framework has better generalization ability especially under longer KB reasoning paths compared to those baselines.

\begin{table*}[!t]
    \centering
    \begin{tabular}{rccccccccc}
         \toprule[1pt]
          & \multicolumn{8}{c}{{\textbf{Movie Domain}}} \\
          \midrule
          & \multicolumn{2}{c}{\textbf{1-Hop}} & \multicolumn{2}{c}{\textbf{2-Hop}} & \multicolumn{2}{c}{\textbf{3-Hop}} & \multicolumn{2}{c}{\textbf{All}}  \\
         \midrule
         & BLEU & F1 & BLEU & F1 & BLEU & F1 & BLEU & F1  \\
         \midrule
         Mem2Seq & 25.6 & 68.9 & 21.8 & 60.8 & 19.5 & 49.2 & 23.9 & 62.0  \\
         GLMP & 30.1 & 77.2 & 28.7 & 72.9 & 27.1 &  61.5 & 28.3 & 73.2   \\
         GraphDialog & 29.2 & 76.6 & 25.6 & 69.1 & 24.7 &  60.6 & 27.2 & 71.6  \\
         DF-Net & 30.6 & 77.4 & 29.5 & 71.6 & 28.9 & 62.1 & 30.3 &  73.5  \\
         \midrule
         \textbf{Ours (Full model)} & \textbf{33.2} & \textbf{82.6} & \textbf{31.3} & \textbf{80.4} & \textbf{30.7} & \textbf{74.9} & \textbf{32.7} &  \textbf{80.6}  \\
         \bottomrule[1pt]
    \end{tabular}
    \caption{Results on the movie domain.}
    \label{movie_result}
\end{table*}

\begin{table*}[!t]
    \centering
    \begin{tabular}{rccccccccc}
         \toprule[1pt]
          & \multicolumn{8}{c}{{\textbf{Hotel Domain}}} \\
          \midrule
          & \multicolumn{2}{c}{\textbf{1-Hop}} & \multicolumn{2}{c}{\textbf{2-Hop}} & \multicolumn{2}{c}{\textbf{3-Hop}} & \multicolumn{2}{c}{\textbf{All}}  \\
         \midrule
         & BLEU & F1 & BLEU & F1 & BLEU & F1 & BLEU & F1  \\
         \midrule
         Mem2Seq & 14.4 & 79.8 & 13.1 & 71.2 & 11.4 & 68.6 & 13.2 & 75.4  \\
         GLMP & 21.3 & 85.5 & 19.8 & 79.4 & 18.9 & 76.2  & 21.0 &  82.9  \\
         GraphDialog & 20.6 & 83.8 & 19.1 & 78.8 & 18.8 & 75.9  & 19.3 & 81.0  \\
         DF-Net & 22.1 & 86.7 & 19.9 & 80.2 & 19.5 & 76.8 & 21.5 &  83.2  \\
         \midrule
         \textbf{Ours (Full model)} & \textbf{23.3} & \textbf{92.4} & \textbf{21.3} & \textbf{89.6} & \textbf{20.7} & \textbf{87.8} & \textbf{22.1} &  \textbf{91.6}  \\
         \bottomrule[1pt]
    \end{tabular}
    \caption{Results on the hotel domain.}
    \label{hotel_result}
\end{table*}

\begin{table*}[!t]
    \centering
    \begin{tabular}{rccccccccc}
         \toprule[1pt]
          & \multicolumn{8}{c}{{\textbf{Restaurant Domain}}} \\
          \midrule
          & \multicolumn{2}{c}{\textbf{1-Hop}} & \multicolumn{2}{c}{\textbf{2-Hop}} & \multicolumn{2}{c}{\textbf{3-Hop}} & \multicolumn{2}{c}{\textbf{All}}  \\
         \midrule
         & BLEU & F1 & BLEU & F1 & BLEU & F1 & BLEU & F1  \\
         \midrule
         Mem2Seq  & 19.0 & 79.8 & 17.3 & 69.4 & 12.4 & 66.3 & 17.0 & 73.7  \\
         GLMP & 22.0 & 90.4 & 19.1 & 83.7 & 18.4 & 80.4  & 20.9 &  86.1  \\
         GraphDialog & 23.2 & 89.9 & 21.2 & 82.1 & 20.6 & 79.8  & 21.4 & 85.0  \\
         DF-Net & 24.5 & 91.5 & 23.0 & 84.2 & 21.1 & 81.0 & 23.3 & 87.3   \\
         \midrule
         \textbf{Ours (Full model)} & \textbf{26.8} & \textbf{96.7} & \textbf{24.4} & \textbf{93.1} & \textbf{22.7} & \textbf{92.2} & \textbf{25.1} & \textbf{94.2}   \\
         \bottomrule[1pt]
    \end{tabular}
    \caption{Results on the restaurant domain.}
    \label{restaurant_result}
\end{table*}
% \begin{table*}[!t]
% \setlength{\belowcaptionskip}{-5mm}
% % \setlength{\abovecaptionskip}{-1mm}
%     \centering
%     \small
%     \begin{tabular}{l|cccc|cccc|cccc}
%          \hline
%           & \multicolumn{4}{c}{\textbf{Movie}} &  \multicolumn{4}{c}{\textbf{Hotel}} &  \multicolumn{4}{c}{\textbf{Restaurant}}  \\
%           \hline
%           & \textbf{1-Hop} & \textbf{2-Hop} & \textbf{3-Hop} & \textbf{All} & \textbf{1-Hop} & \textbf{2-Hop} & \textbf{3-Hop} & \textbf{All} & \textbf{1-Hop} & \textbf{2-Hop} & \textbf{3-Hop} & \textbf{All}  \\
%          \hline
%          Mem2Seq &  &  &  &  &  &  &  &  &  \\
%          GLMP &  &  &  &  &  &   &  &  &  \\
%          GraphDialog &  &  &  &  &  &   &  &  &  \\
%          DF-Net &  &  &  &  &  &  &  &  &  \\
%          \hline
%          \textbf{Ours (Full model)} &  &  &  &  &  &  &  &  &  \\
%          \hline
%     \end{tabular}
%     \caption{Detailed statistics of the synthetic dataset with respect to the number of hops needed by KB reasoning. Numbers in the table are the number of dialogue turns for each category.}
%     \label{dataset_statistics}
% \end{table*}

\subsection{Example Outputs}
We show the generated hypotheses and the proof trees in our framework as shown in Table \ref{case_study_1} and Figure \ref{example_dialogue_1}. As we can see, our model can successfully obtain the correct entities from the KB. Moreover, our framework can formulate sensible hypotheses and generate reasonable proof procedures which can help us gain some insights about the inner workings of our model.

\begin{table*}[!t]
    \centering
    \begin{tabular}{ccccc}
         \toprule[1pt]
         \textbf{Structure} & \textbf{State} & \textbf{Top-5 Candidate} & \textbf{Generated} & \textbf{Belief} \\
         \textbf{Type} & \textbf{Tokens} & \textbf{Tokens} & \textbf{Hypotheses} & \textbf{Scores} \\
         \midrule
         \multirow{5}{*}{H-Hypothesis} &  & ``\textbf{Oakland}'' & ``\textbf{[Oakland, located\_in, Springfield]}'' & \textbf{1.00}  \\
          & ``located\_in ($k$=0)''  & ``Vaudeville'' & ``[Vaudeville, located\_in, Springfield]'' & 0.15 \\
          &   & ``Brown\_Eyes'' & ``[Brown\_Eyes, located\_in, Springfield]'' & 0.17 \\
          & ``Springfield ($k$=1)'' & ``Shipping\_News'' & ``[Shipping\_News, located\_in, Springfield]'' & 0.13 \\
          &  & ``Coburg'' & ``[Coburg, located\_in, Springfield]'' & 0.04 \\
         \bottomrule[1pt]
    \end{tabular}
    \caption{Example outputs on the movie domain of synthetic dataset. Dialogue history: ``I'm looking for a theatre in the Springfield district.''. Generated response: ``Sure I have found a Oakland for you.''. Detailed model working process when generating ``Oakland'' in the response is shown above.}
    \label{case_study_1}
\end{table*}

\begin{figure*}[!ht]
    \centering
    \includegraphics[width=7.0in]{appendix_example_output_1.pdf}
    \caption{Proof tree generated by HRE module for the highest score hypothesis ``[Oakland, Located\_in, Springfield]'' in Table \ref{case_study_1}. The red parts are the predicted bridge entities and the blue parts are the predicted relations for the sub-hypotheses via neural networks. In this case, the model performs 2-hop reasoning (the two leaf node triples) to find the correct KB entity for generating the response.}
    \label{example_dialogue_1}
\end{figure*}

\section{Details on Error Analysis}
\subsection{Structure Prediction Errors}
Given the dialogue history ``\textit{Find me a nearby shopping mall.}'', the generated response is ``\textit{midtown\_shopping\_center is 3\_miles away.}'' and the ground-truth response is ``\textit{The closest shopping\_center is midtown\_shopping\_center 3\_miles away.}''. The generated hypotheses by HG module at decoding timestep 0 are ``\trp{home, address, 5671\_barringer\_street}, \trp{home, address, 481\_amaranta\_ave}, \trp{home, address, midtown\_shopping\_center}, \trp{home, address, 5\_miles}, \trp{home, address, 1\_miles}''. As we can see, the predicted structure type is an T-Hypothesis with an head entity ``\texttt{home}'' and relation ``\texttt{address}''. However, a reasonable structure type should be an H-Hypothesis such as ``\trp{midtown\_shopping\_center, poi\_type, shopping\_center}'' which the model has mispredicted to a T-Hypothesis.

\subsection{Query State Prediction Errors}
Given the dialogue history ``\textit{What time is dinner tonight?}'', the generated response is ``\textit{Dinner is at 11pm.}'' and the ground-truth response is ``\textit{Your dinner is 5pm tonight.}''. The generated hypotheses in HG module for decoding timestep 3 are ``\trp{dinner, traffic\_info, 11pm}, \trp{dinner, traffic\_info, 7pm}, \trp{dinner, traffic\_info, 3pm}, \trp{dinner, traffic\_info, Monday}, \trp{dinner, traffic\_info, Tuesday}''. As we can see, the SP and QSP modules have successfully predicted a sensible T-Hypothesis with ``\texttt{dinner}'' as head entity. However, the predicted relation ``\texttt{traffic\_info}'' is not reasonable according to context where an expected relation would be relations such as ``time''.

\subsection{Candidates Prediction Errors}
Given the dialogue history ``\textit{What time is my doctor\_appointment?}'', the generated response is ``\textit{Your doctor\_appointment is on Thursday.}'' and the ground-truth response is ``\textit{Your doctor\_appointment is scheduled for Friday at 11am.}''. The generated hypotheses by HG for decoding timestep 4 are ``\trp{doctor\_appointment, date, Thursday}, \trp{doctor\_appointment, date, Monday}, \trp{doctor\_appointment, date, 7pm}, \trp{doctor\_appointment, date, Tuesday}, \trp{doctor\_appointment, date, 2pm}''. As we can see, our framework has correctly predicted a T-Hypothesis and state tokens. However, the predicted top-5 candidate tokens don't include the ground-truth date which is \texttt{Friday}, which might affect the final response generation.

\subsection{Belief Score Prediction Errors}
Given the dialogue history ``\textit{Where is home?}'', the generated response is ``\textit{Home is at 819\_alma\_st.}'' and the ground-truth is ``\textit{Home is where is your heart, at 56\_cadwell\_street.}'' The generated hypotheses for decoding timestep 3 are ``\trp{home, address, 819\_alma\_st}, \trp{home, address, 56\_cadwell\_street}, \trp{home, address, 6\_miles}, \trp{home, address, 611\_ames\_ave}, \trp{home, address, 3\_miles}''. As we can see, our framework has predicted a sensible T-Hypothesis with ``\texttt{home}'' as head entity and ``\texttt{address}'' as relation. Also, the CP module has predicted top-5 candidate tail entities which include the ground-truth \texttt{56\_cadwell\_street}. But the HRE module ranked ``\trp{home, address, 819\_alma\_st}'' highest with a score of 0.78 while the ground-truth one ``\trp{home, address, 56\_cadwell\_street}'' is only ranked the second highest with a score of 0.41, which indicates that there is still room for improvements for the HRE module. We plan to investigate and further improve our framework include SP, QSP, CP and HRE in the future work.

\section{Discussions}
\subsection{Why sampling with Gumbel-Softmax instead of directly applying argmax in Hypothesis Generator and Hierarchical Reasoning Engine modules?}
In our preliminary experiments, we have tried using argmax. However, we quickly discovered that it’s not differentiable which hinders our aim of end-to-end differentiability of the whole system. We tried utilizing REINFORCE (reward is obtained by comparing predicted entities with ground-truth entities) to mitigate this issue. However, we find that the results of using argmax+REINFORCE is worse than using Gumbel-Softmax. By checking the sampled tokens from Gumbel-Softmax, we find that it can generate reasonable tokens (Table 5 in the main paper, state tokens etc.), since we have set the temperature parameter of Gumbel-Softmax to 0.1 which approximates argmax largely.

\subsection{Why not expanding the KB using KB completion methods and then use semantic parsing to KB?}
In this work, we are interested in developing an end-to-end trainable framework with explainable KB reasoning. Semantic parsing is one possible alternative. However, when adapting to our own dataset, it requires further annotations for fine-tuning which is costly and time-consuming, and might be not feasible for large-scale datasets. Also, it might induce error propagation issue since the different modules (KB completion, semantic parsing, dialogue encoding and response generation etc.) are not jointly learnt, which can be mitigated by end-to-end approaches. So, in this work, we would like to explore end-to-end approaches for explainable KB reasoning in dialogue response generation.